\documentclass[10pt,twocolumn,letterpaper]{article}

\usepackage{cvpr}      

\makeatletter
\@namedef{ver@everyshi.sty}{}
\makeatother

\usepackage[dvipsnames,svgnames,x11names]{xcolor}
\usepackage{tikz}
\usetikzlibrary{arrows.meta,shapes,calc,matrix,fit,positioning,backgrounds,decorations.markings}
\usepackage{pgfplots}
\usepackage{pgfplotstable}
\pgfplotsset{compat=1.9}
\usepackage{xstring}

\usepgfplotslibrary{external}

\IfBeginWith*{\jobname}{fig/extern/}{\finalcopy}{}

\tikzstyle{every picture}+=[
	remember picture,
	every text node part/.style={align=center},
	every matrix/.append style={ampersand replacement=\&},
]
\tikzstyle{tight} = [inner sep=0pt,outer sep=0pt]
\tikzstyle{node}  = [draw,circle,tight,minimum size=12pt,anchor=center]
\tikzstyle{op}    = [draw,circle,tight]
\tikzstyle{dot}   = [fill,draw,circle,inner sep=1pt,outer sep=0]
\tikzstyle{pt}    = [fill,draw,circle,inner sep=1.5pt,outer sep=.2pt]
\tikzstyle{box}   = [draw,thick,rectangle,inner sep=3pt]
\tikzstyle{high}  = [black!60]
\tikzstyle{group} = [high,box,opacity=.5]
\tikzstyle{rectc} = [tight,transform shape]
\tikzstyle{rect}  = [rectc,anchor=south west]

\tikzset{every mark/.append style={solid}}
\pgfplotsset{
	grid=both, width=\columnwidth, try min ticks=5,
	every axis/.append style={font=\small},
	every axis plot/.append style={thick,mark=none,mark size=1.8,tension=0.18},
	legend cell align=left, legend style={fill opacity=0.8},
	xticklabel={\pgfmathprintnumber[assume math mode=true]{\tick}},
	yticklabel={\pgfmathprintnumber[assume math mode=true]{\tick}},
	nodes near coords math/.style={
		nodes near coords={\pgfmathprintnumber[assume math mode=true]{\pgfplotspointmeta}},
	},
}

\pgfplotsset{
	dash/.style={mark=o,dashed,opacity=0.6},
	dott/.style={mark=o,dotted,opacity=0.6},
	nolim/.style={enlargelimits=false},
	plain/.style={every axis plot/.append style={},nolim,grid=none},
}

\makeatletter
\renewcommand\paragraph{\@startsection{paragraph}{4}{\z@}{1ex}{-1em}{\normalfont\normalsize\bfseries}}
\makeatother

\usepackage{graphicx}
\usepackage{amsmath}
\usepackage{amssymb}
\usepackage[utf8]{inputenc}
\usepackage{mathtools}
\usepackage{enumitem}
\usepackage{xspace}
\usepackage{bm}
\usepackage{array,booktabs}
\usepackage{makecell}
\usepackage{multirow}
\usepackage[numbers,sort&compress]{natbib}
\usepackage[normalem]{ulem}
\usepackage{verbatim}
\usepackage{kantlipsum}
\usepackage{wrapfig}
\usepackage{dsfont}
\usepackage{comment}
\usepackage[accsupp]{axessibility}

\usepackage[pagebackref,breaklinks,colorlinks]{hyperref}
\usepackage[ruled,vlined,linesnumbered]{algorithm2e}

\def\cvprPaperID{9927} 
\def\confName{CVPR}
\def\confYear{2022}

\begin{document}

\title{AlignMixup: Improving Representations By Interpolating Aligned Features}
\author{Shashanka Venkataramanan$^1$ \hspace{0.5em} Ewa Kijak$^1$ \hspace{0.5em} Laurent Amsaleg$^1$ \hspace{0.5em} Yannis Avrithis$^2$  \\
$^1$Inria, Univ Rennes, CNRS, IRISA \hspace{0.5em} $^2$ Athena RC
}

\maketitle

\newcommand{\head}[1]{{\smallskip\noindent\textbf{#1}}}
\newcommand{\alert}[1]{{\color{red}{#1}}}
\newcommand{\sm}{\scriptsize}
\newcommand{\eq}[1]{(\ref{eq:#1})}

\newcommand{\Th}[1]{\textsc{#1}}
\newcommand{\mr}[2]{\multirow{#1}{*}{#2}}
\newcommand{\mc}[2]{\multicolumn{#1}{c}{#2}}
\newcommand{\tb}[1]{\textbf{#1}}
\newcommand{\ch}{\checkmark}

\newcommand{\red}[1]{{\color{red}{#1}}}
\newcommand{\blue}[1]{{\color{blue}{#1}}}
\newcommand{\green}[1]{\color{green}{#1}}
\newcommand{\gray}[1]{{\color{gray}{#1}}}

\newcommand{\citeme}[1]{\red{[XX]}}
\newcommand{\refme}[1]{\red{(XX)}}

\newcommand{\fig}[2][1]{\includegraphics[width=#1\linewidth]{fig/#2}}
\newcommand{\figh}[2][1]{\includegraphics[height=#1\linewidth]{fig/#2}}

\newcommand{\tran}{^\top}
\newcommand{\mtran}{^{-\top}}
\newcommand{\zcol}{\mathbf{0}}
\newcommand{\zrow}{\zcol\tran}

\newcommand{\ind}{\mathbbm{1}}
\newcommand{\expect}{\mathbb{E}}
\newcommand{\nat}{\mathbb{N}}
\newcommand{\zahl}{\mathbb{Z}}
\newcommand{\real}{\mathbb{R}}
\newcommand{\proj}{\mathbb{P}}
\newcommand{\prob}{\mathbf{Pr}}
\newcommand{\normal}{\mathcal{N}}

\newcommand{\mif}{\textrm{if}\ }
\newcommand{\other}{\textrm{otherwise}}
\newcommand{\minimize}{\textrm{minimize}\ }
\newcommand{\maximize}{\textrm{maximize}\ }
\newcommand{\st}{\textrm{subject\ to}\ }

\newcommand{\id}{\operatorname{id}}
\newcommand{\const}{\operatorname{const}}
\newcommand{\sgn}{\operatorname{sgn}}
\newcommand{\var}{\operatorname{Var}}
\newcommand{\mean}{\operatorname{mean}}
\newcommand{\trace}{\operatorname{tr}}
\newcommand{\diag}{\operatorname{diag}}
\newcommand{\vect}{\operatorname{vec}}
\newcommand{\cov}{\operatorname{cov}}
\newcommand{\sign}{\operatorname{sign}}
\newcommand{\prj}{\operatorname{proj}}

\newcommand{\softmax}{\operatorname{softmax}}
\newcommand{\clip}{\operatorname{clip}}

\newcommand{\defn}{\mathrel{:=}}
\newcommand{\peq}{\mathrel{+\!=}}
\newcommand{\meq}{\mathrel{-\!=}}

\newcommand{\floor}[1]{\left\lfloor{#1}\right\rfloor}
\newcommand{\ceil}[1]{\left\lceil{#1}\right\rceil}
\newcommand{\inner}[1]{\left\langle{#1}\right\rangle}
\newcommand{\norm}[1]{\left\|{#1}\right\|}
\newcommand{\abs}[1]{\left|{#1}\right|}
\newcommand{\frob}[1]{\norm{#1}_F}
\newcommand{\card}[1]{\left|{#1}\right|\xspace}
\newcommand{\diff}{\mathrm{d}}
\newcommand{\der}[3][]{\frac{d^{#1}#2}{d#3^{#1}}}
\newcommand{\pder}[3][]{\frac{\partial^{#1}{#2}}{\partial{#3^{#1}}}}
\newcommand{\ipder}[3][]{\partial^{#1}{#2}/\partial{#3^{#1}}}
\newcommand{\dder}[3]{\frac{\partial^2{#1}}{\partial{#2}\partial{#3}}}

\newcommand{\wb}[1]{\overline{#1}}
\newcommand{\wt}[1]{\widetilde{#1}}

\def\xssp{\hspace{1pt}}
\def\ssp{\hspace{3pt}}
\def\msp{\hspace{5pt}}
\def\lsp{\hspace{12pt}}

\newcommand{\cA}{\mathcal{A}}
\newcommand{\cB}{\mathcal{B}}
\newcommand{\cC}{\mathcal{C}}
\newcommand{\cD}{\mathcal{D}}
\newcommand{\cE}{\mathcal{E}}
\newcommand{\cF}{\mathcal{F}}
\newcommand{\cG}{\mathcal{G}}
\newcommand{\cH}{\mathcal{H}}
\newcommand{\cI}{\mathcal{I}}
\newcommand{\cJ}{\mathcal{J}}
\newcommand{\cK}{\mathcal{K}}
\newcommand{\cL}{\mathcal{L}}
\newcommand{\cM}{\mathcal{M}}
\newcommand{\cN}{\mathcal{N}}
\newcommand{\cO}{\mathcal{O}}
\newcommand{\cP}{\mathcal{P}}
\newcommand{\cQ}{\mathcal{Q}}
\newcommand{\cR}{\mathcal{R}}
\newcommand{\cS}{\mathcal{S}}
\newcommand{\cT}{\mathcal{T}}
\newcommand{\cU}{\mathcal{U}}
\newcommand{\cV}{\mathcal{V}}
\newcommand{\cW}{\mathcal{W}}
\newcommand{\cX}{\mathcal{X}}
\newcommand{\cY}{\mathcal{Y}}
\newcommand{\cZ}{\mathcal{Z}}

\newcommand{\vA}{\mathbf{A}}
\newcommand{\vB}{\mathbf{B}}
\newcommand{\vC}{\mathbf{C}}
\newcommand{\vD}{\mathbf{D}}
\newcommand{\vE}{\mathbf{E}}
\newcommand{\vF}{\mathbf{F}}
\newcommand{\vG}{\mathbf{G}}
\newcommand{\vH}{\mathbf{H}}
\newcommand{\vI}{\mathbf{I}}
\newcommand{\vJ}{\mathbf{J}}
\newcommand{\vK}{\mathbf{K}}
\newcommand{\vL}{\mathbf{L}}
\newcommand{\vM}{\mathbf{M}}
\newcommand{\vN}{\mathbf{N}}
\newcommand{\vO}{\mathbf{O}}
\newcommand{\vP}{\mathbf{P}}
\newcommand{\vQ}{\mathbf{Q}}
\newcommand{\vR}{\mathbf{R}}
\newcommand{\vS}{\mathbf{S}}
\newcommand{\vT}{\mathbf{T}}
\newcommand{\vU}{\mathbf{U}}
\newcommand{\vV}{\mathbf{V}}
\newcommand{\vW}{\mathbf{W}}
\newcommand{\vX}{\mathbf{X}}
\newcommand{\vY}{\mathbf{Y}}
\newcommand{\vZ}{\mathbf{Z}}

\newcommand{\va}{\mathbf{a}}
\newcommand{\vb}{\mathbf{b}}
\newcommand{\vc}{\mathbf{c}}
\newcommand{\vd}{\mathbf{d}}
\newcommand{\ve}{\mathbf{e}}
\newcommand{\vf}{\mathbf{f}}
\newcommand{\vg}{\mathbf{g}}
\newcommand{\vh}{\mathbf{h}}
\newcommand{\vi}{\mathbf{i}}
\newcommand{\vj}{\mathbf{j}}
\newcommand{\vk}{\mathbf{k}}
\newcommand{\vl}{\mathbf{l}}
\newcommand{\vm}{\mathbf{m}}
\newcommand{\vn}{\mathbf{n}}
\newcommand{\vo}{\mathbf{o}}
\newcommand{\vp}{\mathbf{p}}
\newcommand{\vq}{\mathbf{q}}
\newcommand{\vr}{\mathbf{r}}
\newcommand{\Vs}{\mathbf{s}}
\newcommand{\vt}{\mathbf{t}}
\newcommand{\vu}{\mathbf{u}}
\newcommand{\vv}{\mathbf{v}}
\newcommand{\vw}{\mathbf{w}}
\newcommand{\vx}{\mathbf{x}}
\newcommand{\vy}{\mathbf{y}}
\newcommand{\vz}{\mathbf{z}}

\newcommand{\vone}{\mathbf{1}}
\newcommand{\vzero}{\mathbf{0}}

\newcommand{\valpha}{{\boldsymbol{\alpha}}}
\newcommand{\vbeta}{{\boldsymbol{\beta}}}
\newcommand{\vgamma}{{\boldsymbol{\gamma}}}
\newcommand{\vdelta}{{\boldsymbol{\delta}}}
\newcommand{\vepsilon}{{\boldsymbol{\epsilon}}}
\newcommand{\vzeta}{{\boldsymbol{\zeta}}}
\newcommand{\veta}{{\boldsymbol{\eta}}}
\newcommand{\vtheta}{{\boldsymbol{\theta}}}
\newcommand{\viota}{{\boldsymbol{\iota}}}
\newcommand{\vkappa}{{\boldsymbol{\kappa}}}
\newcommand{\vlambda}{{\boldsymbol{\lambda}}}
\newcommand{\vmu}{{\boldsymbol{\mu}}}
\newcommand{\vnu}{{\boldsymbol{\nu}}}
\newcommand{\vxi}{{\boldsymbol{\xi}}}
\newcommand{\vomikron}{{\boldsymbol{\omikron}}}
\newcommand{\vpi}{{\boldsymbol{\pi}}}
\newcommand{\vrho}{{\boldsymbol{\rho}}}
\newcommand{\vsigma}{{\boldsymbol{\sigma}}}
\newcommand{\vtau}{{\boldsymbol{\tau}}}
\newcommand{\vupsilon}{{\boldsymbol{\upsilon}}}
\newcommand{\vphi}{{\boldsymbol{\phi}}}
\newcommand{\vchi}{{\boldsymbol{\chi}}}
\newcommand{\vpsi}{{\boldsymbol{\psi}}}
\newcommand{\vomega}{{\boldsymbol{\omega}}}

\newcommand{\rLambda}{\mathrm{\Lambda}}
\newcommand{\rSigma}{\mathrm{\Sigma}}

\newcommand{\vLambda}{\bm{\rLambda}}
\newcommand{\vSigma}{\bm{\rSigma}}

\makeatletter
\newcommand*\bdot{\mathpalette\bdot@{.7}}
\newcommand*\bdot@[2]{\mathbin{\vcenter{\hbox{\scalebox{#2}{$\m@th#1\bullet$}}}}}
\makeatother

\makeatletter
\DeclareRobustCommand\onedot{\futurelet\@let@token\@onedot}
\def\@onedot{\ifx\@let@token.\else.\null\fi\xspace}

\def\eg{\emph{e.g}\onedot} \def\Eg{\emph{E.g}\onedot}
\def\ie{\emph{i.e}\onedot} \def\Ie{\emph{I.e}\onedot}
\def\cf{\emph{cf}\onedot} \def\Cf{\emph{Cf}\onedot}
\def\etc{\emph{etc}\onedot} \def\vs{\emph{vs}\onedot}
\def\wrt{w.r.t\onedot} \def\dof{d.o.f\onedot} \def\aka{a.k.a\onedot}
\def\etal{\emph{et al}\onedot}
\makeatother

\definecolor{ForestGreen}{RGB}{34,139,34}

\newcommand{\mix}{\operatorname{mix}}
\newcommand{\Mix}{\operatorname{Mix}}
\newcommand{\Beta}{\operatorname{Beta}}
\newcommand{\unif}{\operatorname{unif}}

\newcommand{\gp}[1]{\color{ForestGreen}{#1}}  
\newcommand{\gn}[1]{\red{#1}}                 
\newcommand{\se}[1]{\blue{#1}}

\begin{abstract}
Mixup is a powerful data augmentation method that interpolates between two or more examples in the input or feature space and between the corresponding target labels. However, how to best interpolate images is not well defined. Recent mixup methods overlay or cut-and-paste two or more objects into one image, which needs care in selecting regions. Mixup has also been connected to autoencoders, because often autoencoders generate an image that continuously deforms into another. However, such images are typically of low quality.

In this work, we revisit mixup from the \emph{deformation} perspective and introduce AlignMixup, where we geometrically align two images in the feature space. The correspondences allow us to interpolate between two sets of features, while keeping the locations of one set. Interestingly, this retains mostly the geometry or pose of one image and the appearance or texture of the other. We also show that an autoencoder can still improve representation learning under mixup, without the classifier ever seeing decoded images. AlignMixup outperforms state-of-the-art mixup methods on five different benchmarks. Code available at \url{https://github.com/shashankvkt/AlignMixup_CVPR22.git}

\end{abstract}

\section{Introduction}
\label{sec:intro}

\emph{Data augmentation}~\citep{krizhevsky2012imagenet, paulin2014transformation, cubuk2019autoaugment} is a powerful regularization method that increases the amount and diversity of data, be it labeled or unlabeled~\citep{dosovitskiy2013unsupervised}. It improves the generalization performance and helps learning invariance~\citep{simard1998transformation} at almost no cost, because the same example can be transformed in different ways over epochs. However, by operating on one image at a time and limiting to label-preserving transformations, it has limited chances of exploring beyond the image manifold. Hence, it is of little help in combating memorization of training data~\citep{DBLP:conf/iclr/ZhangBHRV17} and sensitivity to adversarial examples~\citep{szegedy2013intriguing}.

\begin{figure}[!h]
\centering
\footnotesize
\setlength{\tabcolsep}{3.0pt}
\newcommand{\sz}{.30}
\newcommand{\mixup}[1]{teaser/#1/#1_28_interpol__4}
\vspace{-10pt}
\begin{tabular}{ccc}
	\fig[\sz]{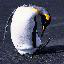} &
	\fig[\sz]{\mixup{IM}} &
	\fig[\sz]{\mixup{cutmix}} \\
	Image 1 &
	Input mixup~\cite{zhang2018mixup} &
	CutMix~\cite{yun2019cutmix} \\[3pt]
	\fig[\sz]{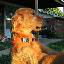} &
	\fig[\sz]{\mixup{mmix}} &
	\fig[\sz]{\mixup{alignmix}} \\
	Image 2 &
	Manifold mixup~\cite{verma2019manifold} &
	AlignMixup (Ours) \\
\end{tabular}
\caption{Different mixup methods. AlignMixup retains the pose of image 2 and the texture of image 1. This  different from overlay (Input and Manifold mixup) or combination of two objects (CutMix). Manifold mixup and AlignMixup
visualized by a decoder (\autoref{sec:viz}) that is not used at training.}
\label{fig:teaser}
\end{figure}

\emph{Mixup} operates on two or more examples at a time, \emph{interpolating} between them in the input space~\citep{zhang2018mixup} or feature space~\citep{verma2019manifold}, while also interpolating between target labels for image classification. This flattens class representations~\citep{verma2019manifold}, reduces overly confident incorrect predictions, and smoothens decision boundaries far away from training data. However, input mixup images are overlays and tend to be unnatural~\citep{yun2019cutmix}. Interestingly, recent mixup methods focus of combining two~\citep{yun2019cutmix, kim2020puzzle} or more~\citep{kim2021co} objects from different images into one in the input space, making efficient use of training pixels. However, randomness in the patch selection and thereby label mixing may mislead the classifier to learn uninformative features ~\citep{uddin2020saliencymix}, which raises the question: \emph{what is a good interpolation of images?}

Bengio \emph{et al.} \cite{bengio2013better} show that traversing along the manifold of representations obtained from deeper layers of the network more likely results in finding realistic examples. This is because the interpolated points smoothly traverse the underlying manifold of the data, capturing salient characteristics of the two images. Furthermore, \cite{berthelot2018understanding} show the ability of autoencoders to capture semantic correspondences obtained by decoding mixed latent codes. This is because the autoencoder may disentangle the underlying factors of variation. Efforts have followed on mixing latent representations of autoencoders to generate realistic images for data augmentation. However, these approaches are more expensive, requiring three networks (encoder, decoder, classifier)~\citep{berthelot2018understanding} and more complex, often also requiring an adversarial discriminator~\citep{beckham2019adversarial, liu2018data}. More importantly, they perform poorly compared to standard input mixup on large datasets~\citep{liu2018data}, due to the low quality of generated images.

In this work, we are motivated by the idea of \emph{deformation} as a natural way of interpolating images, where one image may deform into another, in a continuous way. Contrary to previous efforts, we do not interpolate directly in the input space, we do not limit to vectors as latent codes and we do not decode. We rather investigate geometric \emph{alignment} for mixup, based on explicit semantic correspondences in the feature space. In particular, we explicitly align the feature tensors of two images, resulting in soft correspondences. The tensors can be seen as sets of features with coordinates. Hence, each feature in one set can be interpolated with few features in the other.

By choosing to keep the coordinates of one set or the other, we define an \emph{asymmetric} operation. What we obtain is one object continuously morphing, rather than two objects in one image. Interestingly, observing this asymmetric morphing reveals that we retain the \emph{geometry} or \emph{pose} of the image where we keep the coordinates and the \emph{appearance} or \emph{texture} of the other. \autoref{fig:teaser} illustrates that our method, \emph{AlignMixup}, retains the \emph{pose} of image $2$ and the \emph{texture} of image $1$, which is different from existing mixup methods. Note that, as in manifold mixup, we \emph{do not} decode, hence we are not concerned about the quality of generated images.

We make the following contributions:

\begin{enumerate}[itemsep=1pt, parsep=0pt, topsep=0pt]
	\item We introduce a novel mixup operation, called \emph{AlignMixup}, advocating interpolation of local structure in the feature space (\autoref{sec:align}). Feature tensors are ideal for alignment, giving rise to semantic correspondences and being of low resolution. Alignment is efficient by using \emph{Sinkhorn distance}~\citep{cuturi2013sinkhorn}.
	\item We also show that a \emph{vanilla autoencoder} can further improve representation learning under mixup training, without the classifier seeing decoded clean or mixed images (\autoref{sec:exp}).
	\item We set a new state-of-the-art on \emph{image classification}, \emph{robustness to adversarial attacks}, \emph{calibration}, \emph{weakly-supervised localization} and \emph{out-of-distribution detection} against more sophisticated mixup operations on several networks and datasets (\autoref{sec:exp}).
\end{enumerate}

\section{Related Work}
\label{sec:related}

\paragraph{Mixup}
\cite{zhang2018mixup}, concurrently with similar methods~\citep{inoue2018data,tokozume2017learning}, introduce \emph{mixup}, augmenting data by linear interpolation between two examples. While~\citep{zhang2018mixup} 
apply mixup on intermediate representations, 
it is~\cite{verma2019manifold} who make this work, introducing \emph{manifold mixup}. Without alignment, the result is an overlay of either images~\citep{zhang2018mixup} or features~\citep{verma2019manifold}.~\cite{guo2019mixup} eliminate ``manifold intrusion''---mixed data conflicting with true data. Unlike manifold mixup, AlignMixup interpolates feature tensors from deeper layers after aligning them.

Nonlinear mixing over random image regions is an alternative, \eg from masking square regions~\citep{devries2017improved} to cutting a rectangular region from one image and pasting it onto another~\citep{yun2019cutmix}, as well as several variants using arbitrary regions~\citep{TaMU18,summers2019improved,harris2020fmix}. Instead of choosing regions at random, \emph{saliency} can be used to locate objects from different images and fit them in one~\citep{uddin2020saliencymix,qin2020resizemix,kim2020puzzle,kim2021co}. Exploiting the knowledge of a teacher network to mix images based on saliency has been proposed in ~\citep{dabouei2021supermix}. Instead of combining more than one objects in an image, AlignMixup attempts to deform one object into another.

Another alternative is Automix~\citep{zhu2020automix}, which employs a U-Net rather than an autoencoder, mixing at several layers. It is limited to small datasets and provides little improvement over manifold mixup~\citep{verma2019manifold}. StyleMix and StyleCutMix~\citep{hong2021stylemix} interpolate content and style between two images, using AdaIN~\citep{huang2017arbitrary}, a style transfer autoencoder network. By contrast, AlignMixup aligns feature tensors and interpolates matching features directly, without using any additional network.

\paragraph{Alignment}

Local correspondences from intra-class alignment of feature tensors have been used in \emph{image registration}~\citep{choy2016universal, long2014convnets}, \emph{optical flow}~\citep{weinzaepfel2013deepflow}, \emph{semantic alignment}~\citep{rocco2018end, han2017scnet} and \emph{image retrieval}~\citep{simeoni2019local}. Here, we mostly use \emph{inter-class} alignment. In \emph{few-shot learning}, local correspondences between query and support images are important in finding attention maps, used \eg by CrossTransformers~\citep{doersch2020crosstransformers} and DeepEMD~\citep{zhang2020deepemd}. The \emph{earth mover's distance} (EMD)~\citep{rubner2000earth}, or \emph{Wasserstein metric}, is an instance of \emph{optimal transport}~\citep{villani2008optimal}, addressed by linear programming. To accelerate,~\cite{cuturi2013sinkhorn} computes optimal matching by \emph{Sinkhorn distance} with \emph{entropic regularization}. This distance is widely applied between distributions in generative models~\citep{genevay2018learning, patrini2020sinkhorn}.

EMD has been used for mixup in the input space, for instance \emph{point mixup} for 3D point clouds~\citep{chen2020pointmixup} and OptTransMix for images~\citep{zhu2020automix}, which is the closest to our work. However, aligning coordinates only applies to images with clean background. We rather \emph{align tensors in the feature space}, which is generic. We do so using the Sinkhorn distance, which is orders of magnitude faster than EMD~\citep{cuturi2013sinkhorn}.
\section{AlignMixup}
\label{sec:method}

\subsection{Preliminaries}
\label{sec:prelim}

\paragraph{Problem formulation}

Let $(x,y)$ be an image $x \in \cX$ with its one-hot encoded class label $y \in Y$, where $\cX$ is the input image space, $Y = [0,1]^k$ and $k$ is the number of classes. An \emph{encoder network} $F: \cX \to \real^{c \times w \times h}$ maps $x$ to feature tensor $\vA = F(x)$, where $c$ is the number of channels and $w \times h$ is the spatial resolution.  A \emph{classifier} $g: \real^{c \times w \times h} \to \real^k$ then maps $\vA$ to the vector $p = g(\vA)$ of probabilities over classes.

\paragraph{Mixup}

We follow~\citep{verma2019manifold} in mixing the representations from different layers of the network, focusing on the deepest layers near the classifier. We are given two labeled images $(x, y), (x', y') \in \cX \times Y$. We draw an \emph{interpolation factor} $\lambda \in [0,1]$ from $\Beta(\alpha, \alpha)$~\citep{zhang2018mixup} and then we interpolate labels $y, y'$ linearly by the \emph{standard} mixup operator
\begin{equation}
	\mix_\lambda(y, y') \defn \lambda y + (1 - \lambda) y'
\label{eq:mix-std}
\end{equation}
and inputs $x, x'$ by the generic formula
\begin{equation}
	\Mix_\lambda^{f_1,f_2} (x,x') \defn f_2(\Mix_\lambda(f_1(x), f_1(x')),
\label{eq:mix-in}
\end{equation}
where $\Mix_\lambda$ is a mixup operator to be defined. This generic formula allows interpolation of the input or feature  as $f_2 \circ f_1$ according to
\begin{align}
	\text{input}      \ (x):   & \quad f_1 \defn \id,       f_2 \defn F  \label{eq:lay-x} \\
	\text{feature}    \ (\vA): & \quad f_1 \defn F,         f_2 \defn \id, \label{eq:lay-A}
\end{align}
where $\id$ is the identity mapping. For~\eq{lay-x}, we define $\Mix_\lambda$ in~\eq{mix-in} as standard mixup $\mix_\lambda$~\eq{mix-std}, like~\citep{zhang2018mixup}; while for~\eq{lay-A}, we define $\Mix_\lambda$ as discussed in~\autoref{sec:align}.

By default, we train the encoder network and the classifier by using a classification loss $L_c$ on the output of the classifier $g$ for mixed examples along with the corresponding mixed labels:
\begin{equation}
	L_c(g(\Mix_\lambda^{f_1,f_2} (x,x')), \mix_\lambda(y, y')),
\label{eq:mixed}
\end{equation}
where $L_c(p,y) \defn -\sum_{i=1}^k y_i \log p_i$ is the standard cross-entropy loss. More options using an autoencoder architecture are investigated in \autoref{sec:exp}.

\subsection{Interpolation of aligned feature tensors}
\label{sec:align}

\paragraph{Alignment}

Alignment refers to finding a geometric correspondence between image elements before interpolation. The feature tensor is ideal for this purpose, because its spatial resolution is low, reducing the optimization cost, and allows for semantic correspondence, because features close to the classifier are small.
Importantly, we are not attempting to combine two or more objects into one image~\citep{kim2020puzzle}, but put two objects in correspondence and then interpolate into one. We make no assumptions on the structure of input images in terms of objects and we use no ground truth correspondences.

Our feature tensor alignment is based on \emph{optimal transport} theory~\citep{villani2008optimal} and \emph{Sinkhorn distance} (SD)~\citep{cuturi2013sinkhorn} in particular. Let $\vA \defn F(x), \vA' \defn F(x')$ be the $c \times w \times h$ feature tensors of images $x, x' \in \cX$. We reshape them to $c \times r$ matrices $A, A'$ by flattening the spatial dimensions, where $r \defn hw$. Then, every column $a_j, a'_j \in \real^c$ of $A, A'$ for $j = 1,\dots,r$ is a feature vector representing corresponding to a spatial position in the original image $x,x'$. Let $M$ be the $r \times r$ \emph{cost matrix} with its elements being the pairwise distances of these vectors:
\begin{equation}
	m_{ij} \defn \norm{a_i - a'_j}^2
\label{eq:cost}
\end{equation}
for $i,j \in  \{1,\dots,r\}$. We are looking for a \emph{transport plan}, that is, a $r \times r$ matrix $P \in U_r$, where
\begin{equation}
	U_r \defn \{ P \in \real_+^{r \times r}: P \vone = P\tran \vone = \vone/r \}
\label{eq:poly}
\end{equation}
and $\vone$ is an all-ones vector in $\real^r$. That is, $P$ is non-negative with row-wise and column-wise sum $1/r$, representing a joint probability over spatial positions of $\vA,\vA'$ with uniform marginals. It is chosen to minimize the expected pairwise distance of their features, as expressed by the linear cost function $\inner{P, M}$, under an entropic regularizer:
\begin{equation}
	P^* = \arg\min_{P \in U_r} \inner{P, M} - \epsilon H(P),
\label{eq:opt}
\end{equation}
where $H(P) \defn -\sum_{ij} p_{ij} \log p_{ij}$ is the entropy of $P$, $\inner{\cdot,\cdot}$ is Frobenius inner product and $\epsilon$ is a regularization coefficient. The optimal solution $P^*$ is unique and can be found by forming the $r \times r$ \emph{similarity} matrix $e^{-M/\epsilon}$ and then applying the Sinkhorn-Knopp algorithm~\citep{sinkhornknopp}, \ie, iteratively normalizing rows and columns. A small $\epsilon$ leads to sparser $P$, which improves one-to-one matching but makes the optimization harder \citep{alvarez2018gromov}, while a large $\epsilon$ leads to denser $P$, causing more correspondences and poor matching.

\begin{figure*}
\centering
\fig[1]{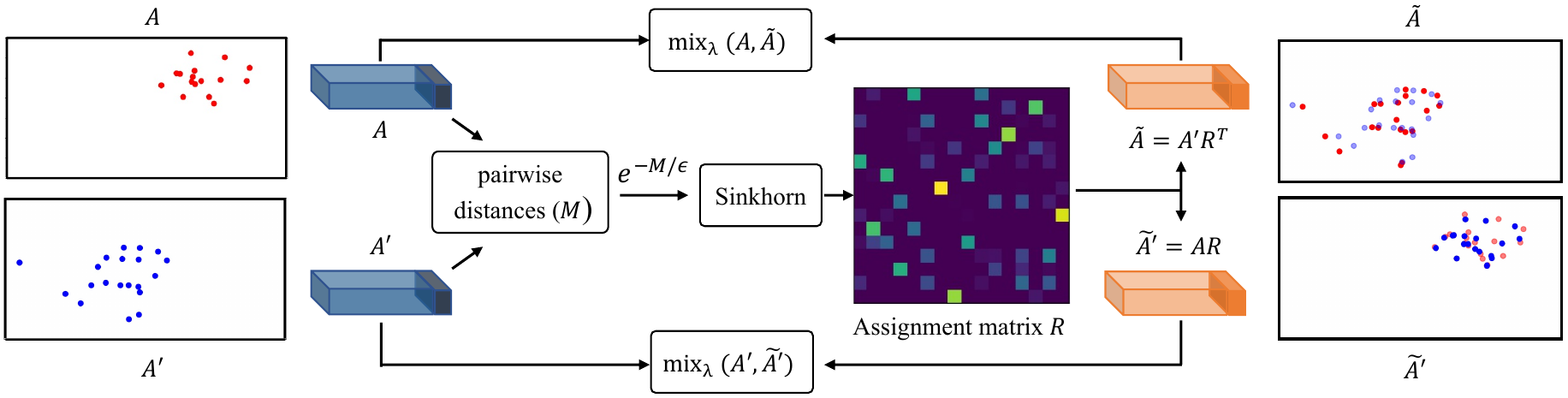}
\caption{\emph{Feature tensor alignment and interpolation}. Cost matrix $M$ contains pairwise distances of feature vectors in tensors $\vA, \vA'$. Assignment matrix $R$ is obtained by Sinkhorn-Knopp~\citep{sinkhornknopp} on similarity matrix $e^{-M/\epsilon}$. $\vA$ is aligned to $\vA'$ according to $R$, giving rise to $\wt{\vA}$. We then interpolate between $\vA, \wt{\vA}$. Symmetrically, we can align $\vA'$ to $\vA$ and interpolate between $\vA', \wt{\vA}'$. $\vA, \vA'$ on the left (toy example of 16 points in 2D) shown semi-transparent on the right for reference.}
\label{fig:align}
\end{figure*}

\paragraph{Interpolation}

The \emph{assignment matrix} $R \defn rP^*$ is a doubly stochastic $r \times r$ matrix whose element $r_{ij}$ expresses the probability that column $a_i$ of $A$ corresponds to column $a'_j$ of $A'$. Thus, we align $A$ and $A'$ as follows:
\begin{align}
	\wt{A}  & \defn A' R\tran \label{eq:extr-1} \\
	\wt{A}' & \defn A  R.     \label{eq:extr-2}
\end{align}
Here, column $\wt{a}_i$ of $c \times r$ matrix $\wt{A}$ is a convex combination of columns of $A'$ that corresponds to the same column $a_i$ of $A$. We reshape $\wt{A}$ back to $c \times w \times h$ tensor $\wt{\vA}$ by expanding spatial dimensions and we say that $\wt{\vA}$ represents $\vA$ \emph{aligned to} $\vA'$. We then interpolate between $\wt{\vA}$ and the original feature tensor $\vA$:
\begin{equation}
    \mix_\lambda(\vA,  \wt{\vA}).
\label{eq:align-1}
\end{equation}
As shown in \autoref{fig:align} (toy example, top right), $\wt{\vA}$ is geometrically close to $\vA$. The correspondence with $\vA'$ and the geometric proximity to $\vA$ makes $\wt{\vA}$ appropriate for interpolation with $\vA$. Symmetrically, we can also \emph{align} $\vA'$ \emph{to} $\vA$ and interpolate between $\wt{\vA}'$ and $\vA'$:
\begin{align}
    \mix_\lambda(\vA', \wt{\vA}').
\label{eq:align-2}
\end{align}
When mixing feature tensors with alignment~\eq{lay-A}, we define $\Mix_\lambda$ in~\eq{mix-in} as the mapping of $(\vA, \vA')$ to either~\eq{align-1} or~\eq{align-2}, chosen at random.

\begin{figure*}
\centering
\small
\setlength{\tabcolsep}{1pt}
\newcommand{\sz}{.058}
\newcommand{\mar}{3pt}
\newcommand{\mixup}[3]{%
	\fig[\sz]{mix/#1/#2/#3/#3_#1_interpol__0} &
	\fig[\sz]{mix/#1/#2/#3/#3_#1_interpol__2} &
	\fig[\sz]{mix/#1/#2/#3/#3_#1_interpol__4} &
	\fig[\sz]{mix/#1/#2/#3/#3_#1_interpol__6} &
	\fig[\sz]{mix/#1/#2/#3/#3_#1_interpol__8} &
	\fig[\sz]{mix/#1/#2/#3/#3_#1_interpol_10} & \hspace{\mar} &
}
\newcommand{\mixlr}[2]{%
	\mixup{#1}{left}{#2}
	\mixup{#1}{right}{#2}
}
\newcommand{\mixfull}[3]{%
	\fig[\sz]{mix/#1/x1} &
	\fig[\sz]{mix/#1/x2} & \hspace{\mar} &
	\mixlr{#1}{zmix}
	$\mix_\lambda(\vA,\vA')$ \\
	\mc{2}{$\mix_\lambda(\vA,\wt{\vA})$} & &
	\mixlr{#1}{alignmix}
	$\mix_\lambda(\vA',\wt{\vA}')$ \\
	& & &
	\mc{6}{(#2)} & \hspace{\mar} &
	\mc{6}{(#3)} & \\[3pt]
}
\begin{tabular}{cccccccccccccccccc}
	$x$ & $x'$ & &
	$0.0$ & $0.2$ & $0.4$ & $0.6$ & $0.8$ & $1.0$ &  &
	$0.0$ & $0.2$ & $0.4$ & $0.6$ & $0.8$ & $1.0$ &  &
	$\lambda$ \\
	\mixfull{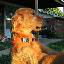}{a}{b}
	\mixfull{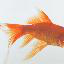}{c}{d}
\end{tabular}
\caption{\emph{Visualizing alignment}. For different $\lambda \in [0,1]$, we interpolate feature tensors  $\vA, \vA'$ without alignment (top) or aligned feature tensors (bottom) of two images $x,x'$ and then we generate a new image by decoding the resulting embedding through the decoder $D$. (a), (c) We align $\vA$ to $\vA'$ and mix with~\eq{align-1}. (b), (d) We align $\vA'$ to $\vA$ and mix with~\eq{align-2}. Only meant for illustration: No decoded images are seen by the classifier at training.}
\label{fig:align_viz}
\end{figure*}

\subsection{Visualization and discussion}
\label{sec:viz}

\paragraph{Decoder}

We use a decoder to study images generated with or without feature alignment. Let $f: \real^{c \times w \times h} \to \real^d$ be a FC layer mapping tensor $\vA$ to embedding $e = f(A)$. We use $f \circ F$ as an encoder and a \emph{decoder} $D: \real^d \to \cX$ mapping $e$ back to the image space, reconstructing image $\hat{x} = D(e)$. The autoencoder is trained using only clean images (without mixup) using \emph{reconstruction} loss $L_r$ between $x$ and $\hat{x}$, where $L_r(x,x') \defn \norm{x-x'}^2$ is the squared Euclidean distance. We use generated images only for visualization purposes below, but we also use the decoder optionally during AlignMixup training in \autoref{sec:exp}.

\paragraph{Discussion}

For different $\lambda \in [0,1]$, we interpolate the feature tensors $\vA, \vA'$ of $x,x'$ without or with alignment, using~\eq{align-1} or~\eq{align-2}, and we generate a new image by decoding the resulting embedding through the decoder $D$.


In \autoref{fig:align_viz}, we visualize such generated images. Interestingly, by aligning $\vA$ to $\vA'$ and mixing using~\eq{align-1} with $\lambda = 0$, the generated image retains the pose of $x$ and the texture of $x'$. In \autoref{fig:align_viz}(a) in particular, when $x$ is `penguin' and $x'$ is `dog', the generated image retains the pose of the penguin, while the texture of the dog aligns to the body of the penguin. Similarly, in \autoref{fig:align_viz}(c), the texture from the goldfish is aligned to that of the stork, while the pose of the stork is retained. Vice versa, as shown in \autoref{fig:align_viz}(b,d), by aligning $\vA'$ to $\vA$ and mixing using~\eq{align-2} with $\lambda = 0$, the generated image retains the pose of $x'$ and the texture of $x$. By contrast, the image generated from unaligned features appears to be an overlay.

Randomly sampling several values of $\lambda \in [0,1]$ during training generates an abundance of samples,  capturing texture from one image and the pose from another. This allows the model to explore beyond the image manifold, thereby improving its generalization and enhancing its performance across multiple benchmarks, as discussed in~\autoref{sec:exp}.

\section{Experiments}
\label{sec:exp}

\subsection{Implementation details}

\paragraph{Architecture}

We use a residual network as our encoder $F$. The output $\vA$ is a $c \times 4 \times 4$ tensor. This is followed by a fully-connected layer as classifier $g$.

\paragraph{Autoencoder}

In \autoref{fig:align_viz}, we have used a decoder to visualize the effect of feature tensor alignment. In our experiments, we also use a decoder optionally during training of AlignMixup, to investigate its effect on representation learning under mixup. This results in a vanilla autoencoder architecture, which we denote as AlignMixup/AE. We use a residual generator~\citep{gulrajani2017improved} as the decoder $D$. The encoder and decoder have the same architecture.

\paragraph{Training}

We train AlignMixup using only the classification loss $L_c$~\eq{mixed} on mixed examples. For a given mini-batch during training, we mix either $x$ or $\vA$ (using either~\eq{align-1} or~\eq{align-2} with alignment). We choose between the three cases uniformly at random. For AlignMixup/AE, we either use the reconstruction loss $L_r$ on clean examples, training the encoder and decoder, or the classification loss $L_c$~\eq{mixed} on mixed examples, training the encoder and classifier. This gives rise to a fourth case and we choose uniformly at random. The algorithm is in the supplementary material.

\paragraph{Hyperparameters} The hyperparameters used for different datasets are reported in the supplementary material.

\begin{table}
\centering
\footnotesize
\setlength{\tabcolsep}{3.5pt}
\begin{tabular}{lccccc} \toprule
	\Th{Dataset}                             & \mc{2}{\Th{Cifar-10}}             & \mc{2}{\Th{Cifar-100}}            & TI              \\
	\Th{Network}                             & R-18            & W16-8           & R-18            & W16-8           & R-18            \\ \midrule
	Baseline                                 & 5.19            & 5.11            & 23.24           & 20.63           & 43.40$^{*}$     \\
	Input~\citep{zhang2018mixup}             & 4.03            & 3.98            & 20.21           & 19.88           & 43.48$^{*}$     \\
	CutMix~\citep{yun2019cutmix}             & 3.27            & 3.54            & 19.37           & 19.71           & 43.11$^{*}$     \\
	Manifold~\citep{verma2019manifold}       & 2.95            & 3.56            & 19.80           & 19.23           & 40.76$^{*}$     \\
	PuzzleMix~\citep{kim2020puzzle}          & 2.93            & \tb{2.99}       & 20.01           & 19.25           & 36.52$^{*}$     \\
	Co-Mixup~\citep{kim2021co}               & \se{2.89}       & \se{3.04}       & 19.81           & 19.57           & 35.85$^{*}$     \\
	SaliencyMix~\citep{uddin2020saliencymix} & 2.99            & 3.53            & 19.69           & 19.59           & 34.81           \\
	StyleMix~\citep{hong2021stylemix}        & 3.76            & 3.89            & 20.04           & 20.45           & 36.13           \\
	StyleCutMix~\citep{hong2021stylemix}     & 3.06            & 3.12            & 19.34           & 19.28           & 34.49           \\ \midrule
	AlignMixup (ours)                        & 2.95            & 3.09            & \se{18.29}      & \se{18.77}      & \se{33.13}      \\
	AlignMixup/AE (ours)                     & \tb{2.83}       & 3.15            & \tb{17.82}      & \tb{18.09}      & \tb{32.73}      \\ \midrule
	Gain                                     & \gp{\tb{+0.06}} & \gn{\tb{-0.10}} & \gp{\tb{+1.52}} & \gp{\tb{+1.14}} & \gp{\tb{+1.76}} \\ \bottomrule
\end{tabular}
\caption{\emph{Image classification} top-1 error (\%) on CIFAR-10/100 and TI (TinyImagenet). Top-1 error (\%): lower is better. Blue: second best. R: PreActResnet, W: WRN. $*$: reported by~\cite{kim2021co}.}
\label{tab:cls}
\end{table}

\begin{table}
\centering
\footnotesize
\setlength{\tabcolsep}{3.5pt}
\begin{tabular}{lcccc} \toprule
	\Th{Method}                                     & \Th{param.}  & \Th{msec/batch}     & \Th{top-1 error}     \\ \midrule
	Baseline                                        & 25M          & 418                 & 23.68            \\
	Input$^\dagger$~\citep{zhang2018mixup}          & 25M          & 436                 & 22.58            \\
	CutMix$^\dagger$~\citep{yun2019cutmix}          & 25M          & 427                 & 21.40            \\
	Manifold$^\dagger$~\citep{verma2019manifold}    & 25M          & 441                 & 22.50            \\
	PuzzleMix$^\dagger$~\citep{kim2020puzzle}       & 25M          & 846                 & 21.24            \\
	Co-Mixup$^{*}$~\citep{kim2021co}                & 25M          & 1022                & --               \\
	SaliencyMix$^{*}$~\citep{uddin2020saliencymix}  & 25M          & 462                 & 21.26            \\
	StyleMix$^{*}$~\citep{hong2021stylemix}         & 25M          & 828                 & -            \\
	StyleCutMix$^{*}$~\citep{hong2021stylemix}      & 25M          & 912                 & -            \\ \midrule
	AlignMixup (ours)                               & 25M          & 450                 & \se{20.68}       \\
	AlignMixup/AE (ours)                            & 35M          & 688                 & \tb{18.83}       \\ \midrule
	Gain                                            &              &                     & \gp{\tb{+2.41}}  \\ \bottomrule
\end{tabular}
\caption{\emph{Image classification} top-1 error (\%) and \emph{computational analysis} on ImageNet using Resnet-50 for 300 epochs. Lower is better. Blue: second best. $*$: reported by authors; $^\dagger$: reported by PuzzleMix. }
\label{tab:cls-imgnet}
\end{table}

\subsection{Image classification and robustness}
\label{sec:exp_class}

We use PreActResnet18~\citep{he2016deep} (R-18) and WRN16-8~\citep{zagoruyko2016wide} as the backbone architecture on CIFAR-10 and CIFAR-100 datasets~\citep{krizhevsky2009learning}. Using the experimental settings of Manifold mixup \cite{verma2019manifold} (in supplementary material), we reproduce the state-of-the-art (SOTA) mixup methods: baseline network (without mixup), Input mixup~\citep{zhang2018mixup}, Manifold mixup~\citep{verma2019manifold}, CutMix~\citep{yun2019cutmix}, PuzzleMix~\citep{kim2020puzzle}, Co-Mixup~\citep{kim2021co}, SaliencyMix~\citep{uddin2020saliencymix}, StyleMix~\citep{hong2021stylemix} and StyleCutMix~\citep{hong2021stylemix} using official code provided by the authors. We do not compare AlignMixup with AutoMix~\citep{zhu2020automix} and Re-Mix~\citep{cao2021remix}, since its experimental settings are different from ours and there is no available code.

In addition, we use R-18 as the backbone network on TinyImagenet~\citep{yao2015tiny} (TI) and reproduce SaliencyMix~\citep{uddin2020saliencymix}, StyleMix~\citep{hong2021stylemix} and StyleCutMix~\citep{hong2021stylemix} following the experimental settings of~\citep{kim2020puzzle}, and Resnet-50 (R-50) on ImageNet~\citep{russakovsky2015imagenet}, following the training protocol of~\citep{kim2020puzzle}. Using top-1 error (\%) as evaluation metric, we show the effectiveness of AlignMixup on image classification and robustness to FGSM~\citep{goodfellow2015explaining} and PGD~\citep{madry2017towards} attacks.

\paragraph{Image classification}

As shown in \autoref{tab:cls}, AlignMixup and AlignMixup/AE is on par or outperforms the SOTA methods by achieving the lowest top-1 error, especially on large datasets. On CIFAR-10, AlignMixup and AlignMixup/AE is on par with Co-Mixup and Puzzlemix with R-18 and WRN16-8. On CIFAR-100, AlignMixup outperforms StyleCutMix and Manifold mixup by 1.05\% and 0.46\% with R-18 and WRN16-8, respectively. On TI, AlignMixup outperforms Co-Mixup by 2.72\% using R-18. From ~\autoref{tab:cls-imgnet}, AlignMixup/AE outperforms PuzzleMix by 2.41\% on ImageNet. While the overall improvement by SOTA methods on ImageNet over Baseline is around 2\%, AlignMixup/AE improves SOTA by another 2.5\%.

\begin{table*}
\centering
\footnotesize
\setlength{\tabcolsep}{6pt}
\begin{tabular}{lccccc|cccc} \toprule
	\Th{Attack}                               & \multicolumn{5}{c}{FGSM}                                                                & \multicolumn{4}{c}{PGD}                                               \\ \midrule
	\Th{Dataset}                              & \mc{2}{\Th{Cifar-10}}             & \mc{2}{\Th{Cifar-100}}            & TI              & \mc{2}{\Th{Cifar-10}}             & \mc{2}{\Th{Cifar-100}}            \\
	\Th{Network}                              & R-18            & W16-8           & R-18            & W16-8           & R-18            & R-18            & W16-8           & R-18            & W16-8           \\ \midrule
	Baseline                                  & 89.41           & 88.02           & 87.12           & 72.81           & 91.85           & 99.99           & 99.94           & 99.97           & 99.99           \\
	Input~\citep{zhang2018mixup}              & 78.42           & 79.21           & 81.30           & 67.33           & 88.68           & 99.77           & 99.43           & 99.96           & 99.37           \\
	CutMix~\citep{yun2019cutmix}              & 77.72           & 78.33           & 86.96           & 60.16           & 88.68           & 99.82           & 98.10           & 98.67           & 97.98           \\
	Manifold~\citep{verma2019manifold}        & 77.63           & 76.11           & 80.29           & 56.45           & 89.25           & 97.22           & 98.49           & 99.66           & 98.43           \\
	PuzzleMix~\citep{kim2020puzzle}           & 57.11           & 60.73           & 78.70           & 57.77           & 83.91           & 97.73           & 97.00           & 96.42           & 95.28           \\
	Co-Mixup~\citep{kim2021co}                & 60.19           & 58.93           & 77.61           & 56.59           & --              & 97.59           & \se{96.19}      & 95.35           & 94.23           \\
	SaliencyMix~\citep{uddin2020saliencymix}  & 57.43           & 68.10           & 77.79           & 58.10           & 81.16           & 97.51           & 97.04           & 95.68           & 93.76           \\
	StyleMix~\citep{hong2021stylemix}         & 79.54           & 71.05           & 80.54           & 67.94           & 84.93           & 98.23           & 97.46           & 98.39           & 98.24           \\
	StyleCutMix~\citep{hong2021stylemix}      & 58.79           & \se{56.12}      & 77.49           & 56.83           & 80.59           & 97.87           & 96.70           & 91.88           & 93.78           \\ \midrule
	AlignMixup (ours)                         & \se{54.83}      & 56.20           & \tb{74.18}      & \tb{55.05}      & \tb{78.83}      & \tb{95.42}      & 96.71           & \tb{90.40}      & \tb{92.16}      \\
	AlignMixup/AE (ours)                      & \tb{52.13}      & \tb{54.86}      & \se{76.40}      & \se{55.44}      & \se{78.98}      & \se{97.16}      & \tb{95.32}      & \se{91.69}      & \se{92.23}      \\ \midrule
	Gain                                      & \gp{\tb{+4.98}} & \gp{\tb{+1.26}} & \gp{\tb{+3.31}} & \gp{\tb{+1.40}} & \gp{\tb{+1.76}} & \gp{\tb{+1.80}} & \gp{\tb{+0.87}} & \gp{\tb{+1.48}} & \gp{\tb{+1.60}} \\ \bottomrule
\end{tabular}
\caption{\emph{Robustness to FGSM \& PGD attacks}. Top-1 error (\%): lower is better. Blue: second best. Gain: reduction of error. TI: TinyImagenet. R: PreActResnet, W: WRN.}
\label{tab:adv}
\end{table*}

\paragraph{Computational complexity}

\autoref{tab:cls-imgnet} shows the computational analysis of AlignMixup training as compared with baseline and SOTA mixup methods on ImageNet, in terms of number of parameters and msec/batch on a NVIDIA RTX 2080 TI GPU. AlignMixup has nearly the same computational overhead as Manifold mixup while achieving 1.82\% increase of accuracy. While SOTA methods like Co-Mixup and PuzzleMix are computationally more expensive than AlignMixup by 1.8$\times$ and 2.3$\times$ respectively, they are outperformed by AlignMixup by 0.6\% on average. AlignMixup/AE brings a further 1.85\% gain in accuracy over AlignMixup. It is important to note that 40\% increase in number of parameters of AlignMixup/AE is due to the residual decoder, which is only used in one out of five cases on clean images without mixup. Computational complexity during inference is the same for all methods.

\paragraph{Challenges}

From~\autoref{tab:cls}, we observe that AlignMixup achieves SoTA top-1 error on CIFAR-10 and CIFAR-100. These results are computed using 2000 epochs following the experimental settings of ~\cite{verma2019manifold}, which also achieves its best performance at 2000 epochs. While baseline mixup methods~\cite{zhang2018mixup, yun2019cutmix, kim2020puzzle, kim2021co, uddin2020saliencymix, hong2021stylemix} perform best at 300 epochs, they do not benefit from long training time. Unlike these methods, which  perform mixup in the image space, Manifold mixup~\cite{verma2019manifold} and AlignMixup performs mixup in the feature space. We hypothesize that this takes longer training time until the network learns some meaningful representations. It is even more challenging in our case, since we mix features at deeper layers comparing with Manifold mixup. Empirically, when trained for 2000 epochs instead of 300 epochs, the top-1 error drops from 21.64 $\rightarrow$ 19.80 for Manifold mixup and from 21.38 $\rightarrow$ 18.29 for AlignMixup.

\paragraph{Robustness to FGSM and PGD attacks}

Following the evaluation protocol of~\citep{kim2020puzzle}, we use $8/255$ $l_{\infty}$ $\epsilon$-ball for FGSM and $4/255$ $l_{\infty}$ $\epsilon$-ball with step size $2/255$ for PGD. We reproduce the results of competitors for FGSM and PGD on CIFAR-10 and CIFAR-100; results of baseline, Input, Manifold, Cutmix and Puzzlemix on TI for FGSM are as reported in~\citep{kim2020puzzle} and reproduced for SaliencyMix, StyleMix and StyleCutMix.

As shown in \autoref{tab:adv}, AlignMixup is more robust comparing to SOTA methods. While AlignMixup is on par with PuzzleMix and Co-Mixup on CIFAR-10 image classification, it outperforms Co-Mixup and PuzzleMix by 5.36\% and 2.28\% in terms of robustness to FGSM attacks. There is also significant gain of robustness to FGSM on Tiny-ImageNet and to the stronger PGD on CIFAR-100.

\begin{table*}
\centering
\footnotesize
\setlength{\tabcolsep}{3pt}
\begin{tabular}{lcccc|cccc|cccc}
\toprule
		\Th{Task}                                 & \mc{11}{\Th{Out-Of-Distribution Detection}}                                                                                                                                                              \\ \midrule
		\Th{Dataset}                              & \mc{4}{\Th{LSUN (crop)}}                                          & \mc{4}{\Th{iSUN}}                                                 & \mc{4}{\Th{TI (crop)}}                                           \\ \midrule
		\mr{2}{\Th{Metric}}                       & \Th{Det}       & \Th{AuROC}     & \Th{AuPR}      & \Th{AuPR}      & \Th{Det}       & \Th{AuROC}     & \Th{AuPR}      & \Th{AuPR}      & \Th{Det}       & \Th{AuROC}     & \Th{AuPR}      & \Th{AuPR}     \\
		                                          & \Th{Acc}       &                & (ID)           & (OOD)          & \Th{Acc}       &                & (ID)           & (OOD)          & \Th{Acc}       &                & (ID)           & (OOD)         \\ \midrule
		Baseline                                  & 54.0           & 47.1           & 54.5           & 45.6           & 66.5           & 72.3           & 74.5           & 69.2           & 61.2           & 64.8           & 67.8           & 60.6         \\
		Input~\citep{zhang2018mixup}              & 57.5           & 59.3           & 61.4           & 55.2           & 59.6           & 63.0           & 60.2           & 63.4           & 58.7           & 62.8           & 63.0           & 62.1         \\
		Cutmix~\citep{yun2019cutmix}              & 63.8           & 63.1           & 61.9           & 63.4           & 67.0           & 76.3           & 81.0           & 77.7           & 70.4           & 84.3           & 87.1           & 80.6         \\
		Manifold~\citep{verma2019manifold}        & 58.9           & 60.3           & 57.8           & 59.5           & 64.7           & 73.1           & 80.7           & 76.0           & 67.4           & 69.9           & 69.3           & 70.5         \\
		PuzzleMix~\citep{kim2020puzzle}           & 64.3           & 69.1           & 80.6           & 73.7           & \se{73.9}      & 77.2           & 79.3           & 71.1           & 71.8           & 76.2           & 78.2           & 81.9         \\
		Co-Mixup~\citep{kim2021co}                & 70.4           & 75.6           & 82.3           & 70.3           & 68.6           & 80.1           & 82.5           & 75.4           & 71.5           & 84.8           & 86.1           & 80.5         \\
		SaliencyMix~\citep{uddin2020saliencymix}  & 68.5           & 79.7           & 82.2           & 64.4           & 65.6           & 76.9           & 78.3           & 79.8           & 73.3           & 83.7           & 87.0           & 82.0         \\
		StyleMix~\citep{hong2021stylemix}         & 62.3           & 64.2           & 70.9           & 63.9           & 61.6           & 68.4           & 67.6           & 60.3           & 67.8           & 73.9           & 71.5           & 78.4         \\
		StyleCutMix~\citep{hong2021stylemix}      & 70.8           & 78.6           & 83.7           & 74.9           & 70.6           & 82.4           & 83.7           & 76.5           & 75.3           & 82.6           & 82.9           & 78.4         \\ \midrule
		AlignMixup (ours)                         & \se{74.2}      & \se{79.9}      & \se{84.1}      & \se{75.1}      & 72.8           & \se{83.2}      & \se{84.1}      & \se{80.3}      & \se{77.2}      & \se{85.0}      & \se{87.8}      & \se{85.0}     \\
		AlignMixup/AE (ours)                      & \tb{76.9}      & \tb{83.5}      & \tb{86.7}      & \tb{79.4}      & \tb{75.6}      & \tb{84.1}      & \tb{85.9}      & \tb{81.7}      & \tb{79.7}      & \tb{88.0}      & \tb{89.7}      & \tb{85.7}     \\ \midrule
		Gain                                      & \gp{\tb{+6.1}} & \gp{\tb{+3.8}} & \gp{\tb{+3.0}} & \gp{\tb{+4.5}} & \gp{\tb{+1.7}} & \gp{\tb{+1.7}} & \gp{\tb{+2.2}} & \gp{\tb{+1.9}} & \gp{\tb{+4.4}} & \gp{\tb{+3.2}} & \gp{\tb{+2.6}} & \gp{\tb{+3.8}} \\ \bottomrule
\end{tabular}
\caption{\emph{Out-of-distribution detection} using PreActResnet18. Det Acc (detection accuracy), AuROC, AuPR (ID) and AuPR (OOD): higher is better; Blue: second best. Gain: increase in performance. TI: TinyImagenet. Additional results are in the supplementary material.}
\label{tab:ood}
\end{table*}

\subsection{Overconfidence}

Deep neural networks tend to be overconfident about incorrect predictions far away from the training data and mixup helps combat this problem. Two standard benchmarks to evaluate this improvement are their ability to detect \emph{out-of-distribution} data and their \emph{calibration}, \ie, the discrepancy between accuracy and confidence.

\paragraph{Out-of-distribution detection}

According to~\citep{hendrycks2016baseline}, \emph{in-distribution} (ID) refers to a test example drawn from the same distribution which the network is trained on, while a sample drawn from any other distribution is \emph{out-of-distribution} (OOD). At inference, given a mixture of ID and OOD examples, the network assigns probabilities to the known classes by softmax. An example is then classified as OOD if the maximum class probability is below a certain threshold, else ID. A well-calibrated network should be able to assign a higher probability to ID than OOD examples, making it easier to distinguish the two distributions.

We compare AlignMixup with SOTA methods trained using R-18 on CIFAR-100 as discussed in \autoref{sec:exp_class}. At inference, ID examples are test images from CIFAR-100, while OOD examples are test images from LSUN (crop)~\citep{yu2015lsun}, iSUN~\citep{xiao2010sun} and Tiny-ImageNet (crop); where crop denotes that the OOD examples are center-cropped to $32\times 32$ to match the resolution of ID images~\citep{yun2019cutmix}. Following~\citep{hendrycks2016baseline}, we measure \emph{detection accuracy} (Det Acc) using a threshold of $0.5$, \emph{area under ROC curve} (AuROC) and \emph{area under precision-recall curve} (AuPR).

As shown in \autoref{tab:ood}, AlignMixup outperforms SOTA methods under all metrics by a large margin, indicating that it is better in reducing over-confident predictions. We further observe that Input mixup is inferior to Baseline, which is consistent with the findings of~\citep{yun2019cutmix}. More results are given in the supplementary material.

\paragraph{Calibration}

According to~\citep{degroot1983comparison}, calibration measures the discrepancy between the accuracy and confidence level of a network's predictions. A poorly calibrated network may make incorrect predictions with high confidence. In the supplementary, we compare AlignMixup with SOTA methods using calibration plots and quantitative experiments.

\begin{table}
\centering
\footnotesize
\setlength{\tabcolsep}{2.5pt}
\begin{tabular}{lcccc} \toprule
    \Th{Metric}                           &\mc{2}{\Th{Top-1 loc.}}           &\mc{2}{\Th{MaxboxAcc-v2}}        \\
	\Th{Network}                          & VGG-GAP        & \Th{ResNet-50} & \Th{VGG-GAP}     & \Th{ResNet-50}  \\ \midrule
	ACoL~\citep{zhang2018adversarial}     & 45.9           & --              & 57.4            & --              \\
	ADL~\citep{choe2019attention}         & 52.4           & --              & 61.3            & 58.4            \\ \midrule
	Baseline CAM~\citep{zhou2016learning} & 37.1           & 49.4            & 59.0            & 59.7            \\
	Input~\citep{zhang2018mixup}          & 41.7           & 49.3            & 57.1            & 60.6            \\
	CutMix~\citep{yun2019cutmix}          & \se{52.5}      & \se{54.8}       & \se{62.6}       & \se{64.8}       \\
	AlignMixup (ours)                     & \tb{53.1}      & \tb{56.2}       & \tb{63.8}       & \tb{65.4}       \\ \midrule
	Gain                                  & \gp{\tb{+0.6}} & \gp{\tb{+1.4}}  & \gp{\tb{+1.2}}  & \gp{\tb{+0.6}}  \\ \bottomrule
\end{tabular}
\caption{\emph{Weakly-supervised object localization} on CUB200-2011. Top-1 loc.: Top-1 localization accuracy (\%), MaxBoxAcc-v2: Maximal box accuracy \citep{choe2020evaluating}. Higher is better. Blue: second best. Gain: increase of accuracy.}
\label{tab:wsol}
\end{table}

\subsection{Weakly-supervised object localization (WSOL)}

WSOL aims to localize an object of interest using only class labels \emph{without bounding boxes} at training. WSOL works by extracting visually discriminative cues to guide the classifier to focus on salient regions in the image.

We train AlignMixup using the same procedure as for image classification. At inference, following~\citep{yun2019cutmix}, we compute a saliency map using CAM~\citep{zhou2016learning}, binarize it using a threshold of $0.15$ and take the bounding box of the mask. We use VGG-GAP~\citep{simonyan2014very} and Resnet-50~\citep{he2016deep} as pretrained on Imagenet~\citep{russakovsky2015imagenet} and we fine-tune them on CUB200-2011~\citep{wah2011caltech}. We follow the evaluation protocol by \cite{choe2020evaluating} and use top-1 localization accuracy with IoU threshold of $0.5$ and Maximal Box Accuracy (MaxBoxAcc-v2) to compare AlignMixup with baseline CAM (without mixup), Input mixup~\citep{zhang2018mixup}, CutOut~\citep{devries2017improved} and CutMix~\citep{yun2019cutmix}.

According to \autoref{tab:wsol}, AlignMixup outperforms Input mixup, CutOut and CutMix by 11.4\%, 7.3\% and 0.6\% respectively using VGG-GAP and by 6.9\%, 3.8\% and 1.4\% respectively using Resnet-50 in terms of top-1 localization accuracy. Furthermore, AlignMixup outperforms CutMix by 1.2\% and 0.6\% using VGG-GAP and Resnet-50 respectively in terms of MaxBoxAcc-v2. It also outperforms dedicated WSOL methods ACoL~\citep{zhang2018adversarial} and ADL~\citep{choe2019attention}, which focus on learning spatially dispersed representations. Qualitative localization results are given in the supplementary material.

\begin{table}
\centering
\footnotesize
\setlength{\tabcolsep}{3pt}
\begin{tabular}{lccc} \toprule
	\Th{Method/Arch}                           & \Th{Layers}                    & \Th{Unaligned} & \Th{Aligned} \\ \midrule
	Baseline                                   &                                & 76.76          & --           \\
	Manifold~\citep{verma2019manifold}         &                                & 80.20          & -            \\
	StyleCutMix~\citep{hong2021stylemix}       &                                & 80.66          & -            \\ \midrule
	\mr{6}{AlignMixup}                         & $\{x, e\}$                     & 80.81          & --           \\
	                                           & $\{\vA\}$                      & 79.07          & 80.28        \\
	                                           & $\{e\}$                        & 78.71          & -            \\
	                                           & $\{x, \vA\}$                   & 80.34          & 81.71        \\
	                                           & $\{x, \vA$, $\vE\}$            & 80.46          & 81.36        \\
	                                           & $\{x, \vA, e\}$                & 80.33          & \tb{81.92}   \\ \midrule
	\mr{6}{AlignMixup/AE}                      & $\{x, e\}$                     & 81.92          & -            \\
	                                           & $\{\vA\}$                      & 79.39          & 81.04        \\
	                                           & $\{e\}$                        & 79.49          & -            \\
	                                           & $\{x, \vA\}$                   & 81.78          & 81.85        \\
	                                           & $\{x, \vE\}$                   & 80.80          & 81.54        \\
	                                           & $\{x, \vA, e\}$                & 81.61          & \tb{82.18}   \\ \midrule
	\mr{3}{AlignMixup/AE}                        & $\{x, \vA_{2 \times 2}, e\}$   & 81.47          & 81.20        \\
	                                           & $\{x, \vA_{4 \times 4}, e\}$   & 81.61          & 82.18        \\
	                                           & $\{x, \vA_{8 \times 8}, e\}$   & 80.49          & \tb{82.20}   \\ \midrule
	\mr{4}{AlignMixup/VAE}                     & $\{x, (\mu, \sigma)\}$         & 81.81          & --           \\
	                                           & $\{x, \vA\}$                   & 81.35          & 81.85        \\
	                                           & $\{x, (\vM, \vSigma)\}$        & 80.45          & 81.10        \\
	                                           & $\{x, \vA, (\mu,\sigma)\}$     & 81.00          & \tb{81.89}   \\ \bottomrule

\end{tabular}
\caption{\emph{Ablations} using R-18 on CIFAR-100. Top-1 classification accuracy (\%): higher is better. Arch: autoencoder architecture. AE: vanilla; VAE: variational~\citep{kingma2013auto}.}
\label{tab:ablation}
\end{table}

\subsection{Ablation study}
\label{sec:ablation}

All ablations are performed on CIFAR-100 using R-18 as encoder $F$ with feature tensor $\vA$ being $512 \times 4 \times 4$. We study the effect of mixing at different layers ($x$, $\vA$), by aligning $\vA$ or not before mixing, as well as using a decoder $D$ in an different autoencoder architectures.  We report top-1 accuracy (\%). All results are in \autoref{tab:ablation}. The ablation showing the effect of the number of iterations in Sinkhorn-Knopp algorithm is summarized in the supplementary material.


\paragraph{Layers}

We study the choice of layers to mix, regardless of feature alignment. According to~\eq{mix-in}, we may mix at any of two layers, represented by $\{x, \vA\}$. To investigate more diverse cases, we introduce an additional layer $f$ to the encoder of AlignMixup and mix its output, which acts as the latent space of AlignMixup/AE. $f$ could be a FC layer, which outputs a vector $e \in \real^{512}$, or a convolutional layer of kernel size $2 \times 2$ and stride $2$, producing a $128 \times 2 \times 2$ tensor $\vE$.
In both cases, mixup is also represented by~\eq{mix-in}, where now $f_1 \defn f \circ F$ and $f_2 \defn \id$. 
Mixing is now chosen from $\{x, \vA, e\}$ or $\{x, \vA, \vE\}$. In AlignMixup (no decoder), among different choices of unaligned layer sets, mixing $\{x, e\}$ results in the highest classification accuracy. Furthermore, AlignMixup/AE outperforms baseline and the best performing competitor StyleCutMix for all choices of layers, even when features are unaligned, showing a motivation to use the decoder.

\paragraph{Tensor alignment}

We investigate the effect of aligning feature tensors or not before mixing it, by using standard mixup~\eq{mix-in} or~\eq{align-1},~\eq{align-2}, respectively. 
It is important to note that when $e$ is a vector, we do not align it. In AlignMixup, we observe that aligning tensors $\vA$ and $\vE$ before mixing improves classification accuracy significantly.
Furthermore, we observe that using an additional FC layer for $e$ brings only minor improvement (81.71 $\rightarrow$ 81.92), meaning that the major improvement comes from alignment. Overall, AlignMixup/AE works the best when $x, \vA, e$ are mixed, with $\vA$ being aligned. It outperforms StyleCutMix by $1.52\%$.

\paragraph{Alignment resolution}

Given the best settings of AlignMixup/AE, we investigate the effect of aligning $\vA$ at different spatial resolutions. The default is $4 \times 4$, denoted as $\vA_{4 \times 4}$. We also experiment $2 \times 2$ ($\vA_{2 \times 2}$), obtained by average pooling, and $8 \times 8$ ($\vA_{8 \times 8}$), by removing downsampling from the last convolutional layer. The accuracy of $8 \times 8$ is only slightly better than $4 \times 4$ by $0.02\%$, while being computationally more expensive. Thus, we choose $4 \times 4$ as the default. By contrast, aligning at $2 \times 2$ is worse than not aligning at all. This may be due to soft correspondences causing loss of information by averaging.

\paragraph{Autoencoder architecture}
We compare AlignMixup with two autoencoder architectures: the \emph{vanilla autoencoder} (AlignMixup/AE), and a \emph{variational autoencoder}~\citep{kingma2013auto} (AlignMixup/VAE).
The latter has two vectors $\mu, \sigma \in \real^{512}$ instead of $e$, representing mean and standard deviation, respectively. We also investigate $128 \times 2 \times 2$ tensors, denoted as $\vM, \vSigma$ where the two variables are mixed simultaneously. As for AlignMixup and AlignMixup/AE, we investigate different combinations of layers with or without alignment. All three architectures work best when $x, \vA, e$ are mixed. Alignment improves consistently on all three architectures. Both AlignMixup and AlignMixup/VAE are inferior to AlignMixup/AE. However, their best settings still outperform Baseline and StyleCutMix. 
\section{Conclusion}
\label{sec:conclusion}

We have shown that mixup of a combination of input and latent representations is a simple and very effective pairwise data augmentation method. The gain is most prominent on large datasets and in combating overconfidence in predictions, as indicated by out-of-distribution detection. Interpolation of feature tensors boosts performance significantly, but only if they are aligned.

Our work is a compromise between a ``good'' hand-crafted interpolation in the image space and a fully learned one in the latent space. A challenge is to make progress in the latter direction without compromising speed and simplicity, which would affect wide applicability.

\section{Acknowledgement}
This work was in part supported by the ANR-19-CE23-0028 MEERQAT project and was performed using the HPC resources from GENCI-IDRIS Grant 2021 AD011012528. This work was partially done while Yannis was at Inria. We also thank Konstantinos Tertikas for his amazing help with adapting AlignMixup to transformers.

{\small
\bibliographystyle{ieee_fullname}
\bibliography{egbib}

\begin{thebibliography}{10}\itemsep=-1pt

\bibitem{alvarez2018gromov}
David Alvarez-Melis and Tommi~S Jaakkola.
\newblock Gromov-wasserstein alignment of word embedding spaces.
\newblock In {\em EMNLP}, 2018.

\bibitem{beckham2019adversarial}
Christopher Beckham, Sina Honari, Vikas Verma, Alex Lamb, Farnoosh Ghadiri,
  R~Devon Hjelm, Yoshua Bengio, and Christopher Pal.
\newblock On adversarial mixup resynthesis.
\newblock In {\em NeurIPS}, 2019.

\bibitem{bengio2013better}
Yoshua Bengio, Gr{\'e}goire Mesnil, Yann Dauphin, and Salah Rifai.
\newblock Better mixing via deep representations.
\newblock In {\em ICML}, 2013.

\bibitem{berthelot2018understanding}
David Berthelot, Colin Raffel, Aurko Roy, and Ian Goodfellow.
\newblock Understanding and improving interpolation in autoencoders via an
  adversarial regularizer.
\newblock In {\em ICLR}, 2019.

\bibitem{cao2021remix}
Jie Cao, Luanxuan Hou, Ming-Hsuan Yang, Ran He, and Zhenan Sun.
\newblock Remix: Towards image-to-image translation with limited data.
\newblock In {\em CVPR}, 2021.

\bibitem{chen2020pointmixup}
Yunlu Chen, Vincent~Tao Hu, Efstratios Gavves, Thomas Mensink, Pascal Mettes,
  Pengwan Yang, and Cees~GM Snoek.
\newblock Pointmixup: Augmentation for point clouds.
\newblock In {\em ECCV}, 2020.

\bibitem{choe2020evaluating}
Junsuk Choe, Seong~Joon Oh, Seungho Lee, Sanghyuk Chun, Zeynep Akata, and
  Hyunjung Shim.
\newblock Evaluating weakly supervised object localization methods right.
\newblock In {\em CVPR}, 2020.

\bibitem{choe2019attention}
Junsuk Choe and Hyunjung Shim.
\newblock Attention-based dropout layer for weakly supervised object
  localization.
\newblock In {\em CVPR}, 2019.

\bibitem{choy2016universal}
Christopher~B Choy, JunYoung Gwak, Silvio Savarese, and Manmohan Chandraker.
\newblock Universal correspondence network.
\newblock In {\em NeurIPS}, 2016.

\bibitem{cubuk2019autoaugment}
Ekin~D Cubuk, Barret Zoph, Dandelion Mane, Vijay Vasudevan, and Quoc~V Le.
\newblock {AutoAugment}: Learning augmentation strategies from data.
\newblock In {\em CVPR}, 2019.

\bibitem{cuturi2013sinkhorn}
Marco Cuturi.
\newblock Sinkhorn distances: lightspeed computation of optimal transport.
\newblock In {\em NeurIPS}, 2013.

\bibitem{dabouei2021supermix}
Ali Dabouei, Sobhan Soleymani, Fariborz Taherkhani, and Nasser~M Nasrabadi.
\newblock Supermix: Supervising the mixing data augmentation.
\newblock In {\em CVPR}, 2021.

\bibitem{degroot1983comparison}
Morris~H DeGroot and Stephen~E Fienberg.
\newblock The comparison and evaluation of forecasters.
\newblock {\em Journal of the Royal Statistical Society: Series D (The
  Statistician)}, 1983.

\bibitem{devries2017improved}
Terrance DeVries and Graham~W Taylor.
\newblock Improved regularization of convolutional neural networks with cutout.
\newblock {\em arXiv preprint arXiv:1708.04552}, 2017.

\bibitem{doersch2020crosstransformers}
Carl Doersch, Ankush Gupta, and Andrew Zisserman.
\newblock Crosstransformers: spatially-aware few-shot transfer.
\newblock In {\em NeurIPS}, 2020.

\bibitem{dosovitskiy2013unsupervised}
Alexey Dosovitskiy, Jost~Tobias Springenberg, and Thomas Brox.
\newblock Unsupervised feature learning by augmenting single images.
\newblock In {\em ICLR Workshops}, 2014.

\bibitem{everingham2010pascal}
Mark Everingham, Luc Van~Gool, Christopher~KI Williams, John Winn, and Andrew
  Zisserman.
\newblock The pascal visual object classes (voc) challenge.
\newblock {\em IJCV}, 2010.

\bibitem{genevay2018learning}
Aude Genevay, Gabriel Peyr{\'e}, and Marco Cuturi.
\newblock Learning generative models with sinkhorn divergences.
\newblock In {\em AISTATS}, 2018.

\bibitem{goodfellow2015explaining}
Ian~J Goodfellow, Jonathon Shlens, and Christian Szegedy.
\newblock Explaining and harnessing adversarial examples.
\newblock In {\em ICLR}, 2015.

\bibitem{Graham_2021_ICCV}
Benjamin Graham, Alaaeldin El-Nouby, Hugo Touvron, Pierre Stock, Armand Joulin,
  Herv\'e J\'egou, and Matthijs Douze.
\newblock Levit: A vision transformer in convnet's clothing for faster
  inference.
\newblock In {\em ICCV}, 2021.

\bibitem{gulrajani2017improved}
Ishaan Gulrajani, Faruk Ahmed, Martin Arjovsky, Vincent Dumoulin, and Aaron
  Courville.
\newblock Improved training of wasserstein gans.
\newblock In {\em ICLR}, 2018.

\bibitem{guo2017calibration}
Chuan Guo, Geoff Pleiss, Yu Sun, and Kilian~Q Weinberger.
\newblock On calibration of modern neural networks.
\newblock In {\em ICML}, 2017.

\bibitem{guo2019mixup}
Hongyu Guo, Yongyi Mao, and Richong Zhang.
\newblock Mixup as locally linear out-of-manifold regularization.
\newblock In {\em AAAI}, 2019.

\bibitem{han2017scnet}
Kai Han, Rafael~S Rezende, Bumsub Ham, Kwan-Yee~K Wong, Minsu Cho, Cordelia
  Schmid, and Jean Ponce.
\newblock Scnet: Learning semantic correspondence.
\newblock In {\em ICCV}, 2017.

\bibitem{harris2020fmix}
Ethan Harris, Antonia Marcu, Matthew Painter, Mahesan Niranjan, and Adam
  Pr{\"u}gel-Bennett~Jonathon Hare.
\newblock Fmix: Enhancing mixed sample data augmentation.
\newblock {\em arXiv preprint arXiv:2002.12047}, 2020.

\bibitem{he2016deep}
Kaiming He, Xiangyu Zhang, Shaoqing Ren, and Jian Sun.
\newblock Deep residual learning for image recognition.
\newblock In {\em CVPR}, 2016.

\bibitem{hendrycks2016baseline}
Dan Hendrycks and Kevin Gimpel.
\newblock A baseline for detecting misclassified and out-of-distribution
  examples in neural networks.
\newblock In {\em ICLR}, 2017.

\bibitem{hong2021stylemix}
Minui Hong, Jinwoo Choi, and Gunhee Kim.
\newblock Stylemix: Separating content and style for enhanced data
  augmentation.
\newblock In {\em CVPR}, 2021.

\bibitem{huang2017arbitrary}
Xun Huang and Serge Belongie.
\newblock Arbitrary style transfer in real-time with adaptive instance
  normalization.
\newblock In {\em ICCV}, 2017.

\bibitem{inoue2018data}
Hiroshi Inoue.
\newblock Data augmentation by pairing samples for images classification.
\newblock {\em arXiv preprint arXiv:1801.02929}, 2018.

\bibitem{kim2021co}
Jang-Hyun Kim, Wonho Choo, Hosan Jeong, and Hyun~Oh Song.
\newblock Co-mixup: Saliency guided joint mixup with supermodular diversity.
\newblock In {\em ICLR}, 2021.

\bibitem{kim2020puzzle}
Jang-Hyun Kim, Wonho Choo, and Hyun~Oh Song.
\newblock Puzzle mix: Exploiting saliency and local statistics for optimal
  mixup.
\newblock In {\em ICML}, 2020.

\bibitem{kingma2013auto}
Diederik~P Kingma and Max Welling.
\newblock Auto-encoding variational bayes.
\newblock In {\em ICLR}, 2014.

\bibitem{sinkhornknopp}
Philip~A Knight.
\newblock The {Sinkhorn-Knopp} algorithm: convergence and applications.
\newblock {\em SIAM Journal on Matrix Analysis and Applications}, 2008.

\bibitem{krizhevsky2009learning}
Alex Krizhevsky, Geoffrey Hinton, et~al.
\newblock Learning multiple layers of features from tiny images.
\newblock Technical report, University of Toronto, 2009.

\bibitem{krizhevsky2012imagenet}
Alex Krizhevsky, Ilya Sutskever, and Geoffrey~E Hinton.
\newblock Imagenet classification with deep convolutional neural networks.
\newblock In {\em NIPS}, 2012.

\bibitem{lin2014microsoft}
Tsung-Yi Lin, Michael Maire, Serge Belongie, James Hays, Pietro Perona, Deva
  Ramanan, Piotr Doll{\'a}r, and C~Lawrence Zitnick.
\newblock Microsoft coco: Common objects in context.
\newblock In {\em ECCV}, 2014.

\bibitem{liu2016ssd}
Wei Liu, Dragomir Anguelov, Dumitru Erhan, Christian Szegedy, Scott Reed,
  Cheng-Yang Fu, and Alexander~C Berg.
\newblock Ssd: Single shot multibox detector.
\newblock In {\em ECCV}, 2016.

\bibitem{liu2018data}
Xiaofeng Liu, Yang Zou, Lingsheng Kong, Zhihui Diao, Junliang Yan, Jun Wang,
  Site Li, Ping Jia, and Jane You.
\newblock Data augmentation via latent space interpolation for image
  classification.
\newblock In {\em ICPR}, 2018.

\bibitem{long2014convnets}
Jonathan Long, Ning Zhang, and Trevor Darrell.
\newblock Do convnets learn correspondence?
\newblock In {\em NIPS}, 2014.

\bibitem{madry2017towards}
Aleksander Madry, Aleksandar Makelov, Ludwig Schmidt, Dimitris Tsipras, and
  Adrian Vladu.
\newblock Towards deep learning models resistant to adversarial attacks.
\newblock In {\em ICLR}, 2018.

\bibitem{patrini2020sinkhorn}
Giorgio Patrini, Rianne van~den Berg, Patrick Forre, Marcello Carioni, Samarth
  Bhargav, Max Welling, Tim Genewein, and Frank Nielsen.
\newblock Sinkhorn autoencoders.
\newblock In {\em Uncertainty in Artificial Intelligence}, 2020.

\bibitem{paulin2014transformation}
Mattis Paulin, J{\'e}r{\^o}me Revaud, Zaid Harchaoui, Florent Perronnin, and
  Cordelia Schmid.
\newblock Transformation pursuit for image classification.
\newblock In {\em CVPR}, 2014.

\bibitem{qin2020resizemix}
Jie Qin, Jiemin Fang, Qian Zhang, Wenyu Liu, Xingang Wang, and Xinggang Wang.
\newblock Resizemix: Mixing data with preserved object information and true
  labels.
\newblock {\em arXiv preprint arXiv:2012.11101}, 2020.

\bibitem{ren2015faster}
Shaoqing Ren, Kaiming He, Ross Girshick, and Jian Sun.
\newblock Faster r-cnn: Towards real-time object detection with region proposal
  networks.
\newblock In {\em NIPS}, 2015.

\bibitem{rocco2018end}
Ignacio Rocco, Relja Arandjelovi{\'c}, and Josef Sivic.
\newblock End-to-end weakly-supervised semantic alignment.
\newblock In {\em CVPR}, 2018.

\bibitem{rubner2000earth}
Yossi Rubner, Carlo Tomasi, and Leonidas~J Guibas.
\newblock The earth mover's distance as a metric for image retrieval.
\newblock {\em IJCV}, 2000.

\bibitem{russakovsky2015imagenet}
Olga Russakovsky, Jia Deng, Hao Su, Jonathan Krause, Sanjeev Satheesh, Sean Ma,
  Zhiheng Huang, Andrej Karpathy, Aditya Khosla, Michael Bernstein, et~al.
\newblock Imagenet large scale visual recognition challenge.
\newblock {\em IJCV}, 2015.

\bibitem{simard1998transformation}
Patrice~Y Simard, Yann~A LeCun, John~S Denker, and Bernard Victorri.
\newblock Transformation invariance in pattern recognition—tangent distance
  and tangent propagation.
\newblock In {\em Neural networks: tricks of the trade}. 1998.

\bibitem{simeoni2019local}
Oriane Sim{\'e}oni, Yannis Avrithis, and Ondrej Chum.
\newblock Local features and visual words emerge in activations.
\newblock In {\em CVPR}, 2019.

\bibitem{simonyan2014very}
Karen Simonyan and Andrew Zisserman.
\newblock Very deep convolutional networks for large-scale image recognition.
\newblock In {\em ICLR}, 2015.

\bibitem{summers2019improved}
Cecilia Summers and Michael~J Dinneen.
\newblock Improved mixed-example data augmentation.
\newblock In {\em WACV}, 2019.

\bibitem{szegedy2013intriguing}
Christian Szegedy, Wojciech Zaremba, Ilya Sutskever, Joan Bruna, Dumitru Erhan,
  Ian Goodfellow, and Rob Fergus.
\newblock Intriguing properties of neural networks.
\newblock In {\em ICLR}, 2014.

\bibitem{TaMU18}
Ryo Takahashi, Takashi Matsubara, and Kuniaki Uehara.
\newblock Ricap: Random image cropping and patching data augmentation for deep
  cnns.
\newblock In {\em ACML}, 2018.

\bibitem{thulasidasan2019mixup}
Sunil Thulasidasan, Gopinath Chennupati, Jeff Bilmes, Tanmoy Bhattacharya, and
  Sarah Michalak.
\newblock On mixup training: Improved calibration and predictive uncertainty
  for deep neural networks.
\newblock In {\em NeurIPS}, 2019.

\bibitem{tokozume2017learning}
Yuji Tokozume, Yoshitaka Ushiku, and Tatsuya Harada.
\newblock Learning from between-class examples for deep sound recognition.
\newblock In {\em ICLR}, 2018.

\bibitem{uddin2020saliencymix}
A~F~M Uddin, Mst. Monira, Wheemyung Shin, TaeChoong Chung, and Sung-Ho Bae.
\newblock {SaliencyMix}: A saliency guided data augmentation strategy for
  better regularization.
\newblock In {\em ICML}, 2021.

\bibitem{verma2019manifold}
Vikas Verma, Alex Lamb, Christopher Beckham, Amir Najafi, Ioannis Mitliagkas,
  David Lopez-Paz, and Yoshua Bengio.
\newblock Manifold mixup: Better representations by interpolating hidden
  states.
\newblock In {\em ICML}, 2019.

\bibitem{villani2008optimal}
C{\'e}dric Villani.
\newblock {\em Optimal transport: old and new}.
\newblock Springer Science \& Business Media, 2008.

\bibitem{wah2011caltech}
Catherine Wah, Steve Branson, Peter Welinder, Pietro Perona, and Serge
  Belongie.
\newblock The caltech-ucsd birds-200-2011 dataset.
\newblock Technical Report CNS-TR-2011-001, California Institute of Technology,
  2011.

\bibitem{weinzaepfel2013deepflow}
Philippe Weinzaepfel, Jerome Revaud, Zaid Harchaoui, and Cordelia Schmid.
\newblock Deepflow: Large displacement optical flow with deep matching.
\newblock In {\em ICCV}, 2013.

\bibitem{xiao2010sun}
Jianxiong Xiao, James Hays, Krista~A Ehinger, Aude Oliva, and Antonio Torralba.
\newblock Sun database: Large-scale scene recognition from abbey to zoo.
\newblock In {\em CVPR}, 2010.

\bibitem{yao2015tiny}
Leon Yao and John Miller.
\newblock Tiny imagenet classification with convolutional neural networks.
\newblock Technical report, Standford University, 2015.

\bibitem{yu2015lsun}
Fisher Yu, Ari Seff, Yinda Zhang, Shuran Song, Thomas Funkhouser, and Jianxiong
  Xiao.
\newblock Lsun: Construction of a large-scale image dataset using deep learning
  with humans in the loop.
\newblock {\em arXiv preprint arXiv:1506.03365}, 2015.

\bibitem{yun2019cutmix}
Sangdoo Yun, Dongyoon Han, Seong~Joon Oh, Sanghyuk Chun, Junsuk Choe, and
  Youngjoon Yoo.
\newblock Cutmix: Regularization strategy to train strong classifiers with
  localizable features.
\newblock In {\em ICCV}, 2019.

\bibitem{zagoruyko2016wide}
Sergey Zagoruyko and Nikos Komodakis.
\newblock Wide residual networks.
\newblock In {\em BMVC}, 2016.

\bibitem{DBLP:conf/iclr/ZhangBHRV17}
Chiyuan Zhang, Samy Bengio, Moritz Hardt, Benjamin Recht, and Oriol Vinyals.
\newblock Understanding deep learning requires rethinking generalization.
\newblock In {\em ICLR}, 2017.

\bibitem{zhang2020deepemd}
Chi Zhang, Yujun Cai, Guosheng Lin, and Chunhua Shen.
\newblock Deepemd: Few-shot image classification with differentiable earth
  mover's distance and structured classifiers.
\newblock In {\em CVPR}, 2020.

\bibitem{zhang2018mixup}
Hongyi Zhang, Moustapha Cisse, Yann~N Dauphin, and David Lopez-Paz.
\newblock mixup: Beyond empirical risk minimization.
\newblock In {\em ICLR}, 2018.

\bibitem{zhang2018adversarial}
Xiaolin Zhang, Yunchao Wei, Jiashi Feng, Yi Yang, and Thomas~S Huang.
\newblock Adversarial complementary learning for weakly supervised object
  localization.
\newblock In {\em CVPR}, 2018.

\bibitem{zhou2016learning}
Bolei Zhou, Aditya Khosla, Agata Lapedriza, Aude Oliva, and Antonio Torralba.
\newblock Learning deep features for discriminative localization.
\newblock In {\em CVPR}, 2016.

\bibitem{zhu2020automix}
Jianchao Zhu, Liangliang Shi, Junchi Yan, and Hongyuan Zha.
\newblock Automix: Mixup networks for sample interpolation via cooperative
  barycenter learning.
\newblock In {\em ECCV}, 2020.

\end{thebibliography}
}


\clearpage



\appendix

\section{Algorithm}
\label{sec:supp-algorithm}

AlignMixup and AlignMixup/AE are summarized in \autoref{al:AlignMixup}. By default (AlignMixup), for each mini-batch, we uniformly draw at random one among three choices (line~\ref{al:mode}) over mixup on input ($x$) or feature tensors ($\vA$, using either~\eq{align-1} or~\eq{align-2} for mixing). For AlignMixup/AE, there is a fourth choice where we only use reconstruction loss on clean examples (line \ref{al:clean}).

For mixup, we use only classification loss~\eq{mixed} (line~\ref{al:mixed}). Following~\citep{verma2019manifold}, we form, for each example $(x,y)$ in the mini-batch, a paired example $(x',y')$ from the same mini-batch regardless of class labels, by randomly permuting the indices (lines~\ref{al:perm},\ref{al:pair}). Inputs $x,x'$ are mixed by~\eq{mix-in},\eq{lay-x} (line~\ref{al:input}). Feature tensors $\vA$ and $ \vA'$ are first aligned and then mixed by~\eq{mix-in},\eq{align-1} ($\vA$ aligns to $\vA'$) or~\eq{mix-in},\eq{align-2} ($\vA'$ aligns to $\vA$) (lines~\ref{al:swap},\ref{al:feat}).

\begin{algorithm}
\small

\SetFuncSty{textsc}
\SetDataSty{emph}
\newcommand{\commentsty}[1]{{\color{ForestGreen}#1}}
\SetCommentSty{commentsty}
\SetKwComment{Comment}{$\triangleright$ }{}

\SetKwData{mode}{mode}
\SetKwData{clean}{clean}
\SetKwData{input}{input}
\SetKwData{feat}{feat}
\SetKwData{embed}{embed}
\SetKwFunction{Swap}{swap}
\SetKwFunction{Reshape}{reshape}
\SetKwFunction{Dist}{dist}
\SetKwFunction{Sinkhorn}{sinkhorn}
\SetKwFunction{Detach}{detach}

\KwIn{encoders $F$; \blue{embedding $e$, decoder $D$;} classifier $g$}
\KwIn{mini-batch $B \defn \{(x_i, y_i)\}_{i=1}^b$}
\KwOut{loss values $L \defn \{\ell_i\}_{i=1}^b$}

$\pi \sim \unif (S_b)$ \Comment*[f]{random permutation of $\{1,\dots,b\}$} \label{al:perm} \\
$\mode \sim \unif \{ \blue{\clean}, \input, \feat, \feat' \}$ \Comment*[f]{mixup?} \label{al:mode} \\
\For{$i \in \{1,\dots,b\}$}{
	$(x,y) \gets (x_i, y_i)$ \Comment*[f]{current example} \\
	\blue{
	\If(\Comment*[f]{no mixup}){\mode $=$ \clean}{
		$\hat{x} \gets D(e(F(x)))$ \Comment*[f]{encode/decode} \\
		$\ell_i \gets L_r(x, \hat{x})$ \Comment*[f]{reconstruction loss} \label{al:clean}
	}
	}
	\Else(\Comment*[f]{mixup}){
    	$\lambda \sim \Beta(\alpha,\alpha)$ \Comment*[f]{interpolation factor} \\
    	$(x',y') \gets (x_{\pi(i)}, y_{\pi(i)})$ \Comment*[f]{paired example} \label{al:pair} \\
    	\If(\Comment*[f]{as in~\citep{zhang2018mixup}}){$\mode = \input$}{
    		$out \gets F(\mix_\lambda(x, x'))$} \Comment*[f]{\eq{mix-in},\eq{lay-x} \label{al:input}
    	}
    	\Else(\Comment*[f]{$\mode \in \{\feat,\feat'\}$}){
    		\If(\Comment*[f]{choose~\eq{align-2} over~\eq{align-1}}){$\mode = \feat'$}{
    			\Swap($x$, $x'$), \Swap($y$, $y'$) \label{al:swap}
    		}
    		$\vA \gets F(x)$ , $\vA' \gets F(x')$ \Comment*[f]{feature tensors} \\
    		$A \gets$ \Reshape$_{c \times r}$($\vA$) \Comment*[f]{to matrix} \\
    		$A' \gets$ \Reshape$_{c \times r}$($\vA'$)\\
    		$M \gets \Dist(A,A')$ \Comment*[f]{pairwise distances~\eq{cost}}  \\
    		$P^* \gets \Sinkhorn(\exp(-M/\epsilon))$ \Comment*[f]{tran. plan~\eq{opt}} \\
    		$R \gets \Detach(r P^*)$ \Comment*[f]{assignments} \label{al:detach} \\
    		$\wt{A} \gets A' R\tran$ \Comment*[f]{alignment ~\eq{extr-1}} \\
    		$\wt{\vA} \gets$ \Reshape$_{c \times w \times h}$($\wt{A}$) \Comment*[f]{to tensor} \\
    		$out \gets f(\mix_\lambda(\vA, \wt{\vA}))$ \Comment*[f]{\eq{mix-in},\eq{align-1}} \label{al:feat}
    	}
    	$\ell_i \gets L_c(g(out), \mix_\lambda(y, y'))$ \Comment*[f]{classification loss~\eq{mixed}} \label{al:mixed}
	}
}
\caption{AlignMixup\blue{/AE} (parts involved in the AE variant indicated in \blue{blue})}
\label{al:AlignMixup}
\end{algorithm}

In computing loss derivatives, we backpropagate through feature tensors $\vA,\vA'$ but not through the transport plan $P^*$ (line~\ref{al:detach}). Hence, although the Sinkhorn-Knopp algorithm~\citep{sinkhornknopp} is differentiable, its iterations take place only in the forward pass. Importantly, AlignMixup is easy to implement and does not require sophisticated optimization like~\citep{kim2020puzzle, kim2021co}.

\section{Hyperparameter settings}
\label{sec:supp-hyperparams}

\paragraph{CIFAR-10/CIFAR-100}

We train AlignMixup using SGD for $2000$ epochs with an initial learning rate of $0.1$, decayed by a factor $0.1$ every $500$ epochs. We set the momentum as $0.9$ with a weight decay of $0.0001$ and use a batch size of $128$. The interpolation factor is drawn from $\Beta(\alpha, \alpha)$ where $\alpha = 2.0$. Using these settings, we reproduce the results of SOTA mixup methods for image classification, robustness to FGSM and PGD attacks, calibration and out-of-distribution detection. For alignment, we apply the Sinkhorn-Knopp algorithm~\citep{sinkhornknopp} for $100$ iterations with entropic regularization coefficient $\epsilon =0.1$.

\paragraph{TinyImagenet}

We follow the training protocol of Kim~\etal~\citep{kim2020puzzle}, training R-18 as stage-1 encoder $F$ using SGD for $1200$ epochs. We set the initial learning rate to $0.1$ and decay it by $0.1$ at $600$ and $900$ epochs. We set the momentum as $0.9$ with a weight decay of $0.0001$ and use a batch size of $128$ on $2$ GPUs. The interpolation factor is drawn from $\Beta(\alpha, \alpha)$ where $\alpha = 2.0$. For alignment, we apply the Sinkhorn-Knopp algorithm~\citep{sinkhornknopp} for $100$ iterations with entropic regularization coefficient $\epsilon =0.1$.

\paragraph{ImageNet}

We follow the training protocol of Kim~\etal~\citep{kim2020puzzle}, where training R-50 as $F$ using SGD for $300$ epochs. The initial learning rate of the classifier and the remaining layers is set to $0.1$ and $0.01$, respectively. We decay the learning rate by $0.1$ at $100$ and $200$ epochs. We set the momentum as $0.9$ with a weight decay of $0.0001$ and use a batch size of $100$ on $4$ GPUs. The interpolation factor is drawn from $\Beta(\alpha, \alpha)$ where $\alpha = 2.0$. For alignment, we apply the Sinkhorn-Knopp algorithm~\citep{sinkhornknopp} for $100$ iterations with entropic regularization coefficient $\epsilon =0.1$.

We also train R-50 on ImageNet for $100$ epochs, following the training protocol described in Kim~\etal~\citep{kim2021co}.

\paragraph{CUB200-2011}

For weakly-supervised object localization (WSOL), we use VGG-GAP and R-50 pretrained on ImageNet as $F$. The training strategy for WSOL is the same as image classification and the network is trained \emph{without bounding box information}. In R-50,  following~\citep{yun2019cutmix}, we modify the last residual block (\verb+layer 4+) to have stride $2$ instead of $1$, resulting in a feature map of spatial resolution $14 \times 14$. The modified architecture of VGG-GAP is the same as described in~\citep{zhou2016learning}. The classifier is modified to have $200$ classes instead of $1000$.

For fair comparisons with~\citep{yun2019cutmix}, during training, we resize the input image to $256 \times 256$ and randomly crop the resized image to $224 \times 224$. During testing, we directly resize to $224 \times 224$. We train the network for 600 epochs using SGD. For R-50, the initial learning rate of the classifier and the remaining layers is set to $0.01$ and $0.001$, respectively. For VGG, the initial learning rate of the classifier and the remaining layers is set to $0.001$ and $0.0001$, respectively. We decay the learning rate by $0.1$ every 150 epochs. The momentum is set to $0.9$ with weight decay of $0.0001$ and batch size of $16$.

\begin{figure*}[h!]
\centering
\fig[1]{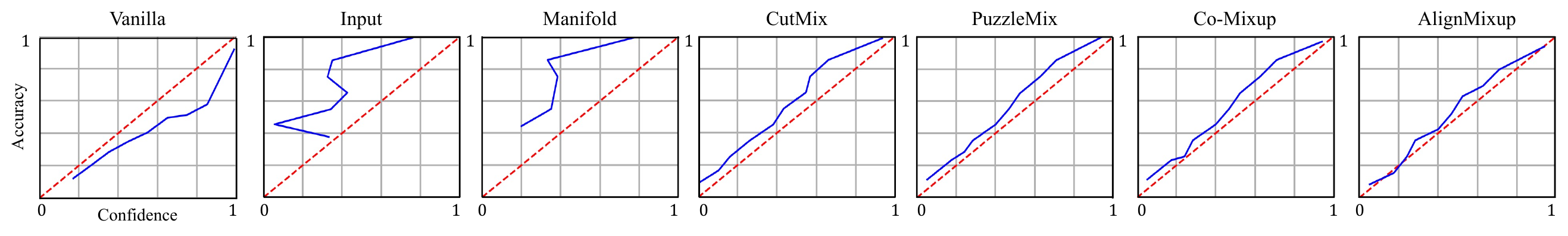}
\caption{\emph{Calibration plots} on CIFAR-100 using PreActResnet18: near diagonal is better. We plot accuracy \vs confidence, that is, probability for the predicted class.}
\label{fig:ece}
\end{figure*}

\section{Additional experiments}
\label{sec:supp-add-exp}
\begin{table}
\centering
\small
\begin{tabular}{lc} \toprule
	\Th{Network}                                & \Th{Resnet-50}   \\ \midrule
	Baseline                                    & 24.03            \\
	Input~\citep{zhang2018mixup}                & 22.97            \\
	Manifold~\citep{verma2019manifold}          & 23.30            \\
	CutMix~\citep{yun2019cutmix}                & 22.92            \\
	PuzzleMix~\citep{kim2020puzzle}             & 22.49            \\
	Co-Mixup~\citep{kim2021co}                  & \se{22.39}       \\
	StyleMix~\citep{hong2021stylemix}           & 24.06            \\
	StyleCutMix~\citep{hong2021stylemix}        & 22.71            \\ \midrule
	AlignMixup (ours)                           & \tb{22.0}       \\ \midrule
	Gain                                        & \gp{\tb{+0.39}}  \\ \bottomrule
\end{tabular}
\caption{\emph{Image classification} on ImageNet for $100$ epochs using ResNet-50. Top-1 error (\%): lower is better. Blue: second best. Gain: reduction of error.}
\label{tab:imgnet}
\end{table}
\begin{table}
\centering
\scriptsize
\setlength{\tabcolsep}{1pt}
\begin{tabular}{lcccc@{\hskip 4pt}|@{\hskip 6pt}cccc} \toprule
		\Th{Dataset}                              & \mc{4}{\Th{LSUN (resize)}}                                        & \mc{4}{\Th{TI (resize)}}                                          \\ \midrule
		\mr{2}{\Th{Metric}}                       & \Th{Det}       & \Th{Au}        & \Th{AuPR}      & \Th{AuPR}      & \Th{Det}       & \Th{Au}        & \Th{AuPR}      & \Th{AuPR}      \\
		                                          & \Th{Acc}       & \Th{ROC}       & (ID)           & (OOD)          & \Th{Acc}       & \Th{ROC}       & (ID)           & (OOD)          \\ \midrule
		Baseline                                  & 67.6           & 73.3           & 76.6           & 68.9           & 65.1           & 70.6           & 73.1           & 67.1           \\
		Input~\citep{zhang2018mixup}              & 61.5           & 66.5           & 66.4           & 65.8           & 59.6           & 63.8           & 63.0           & 63.4           \\
		Cutmix~\citep{yun2019cutmix}              & 71.3           & 77.4           & 79.1           & 75.5           & 69.1           & 79.4           & 79.8           & 73.3           \\
		Manifold~\citep{verma2019manifold}        & 67.8           & 78.9           & 76.3           & 71.3           & 62.5           & 77.8           & 76.8           & 72.2           \\
		PuzzleMix~\citep{kim2020puzzle}           & 74.9           & 79.9           & 84.0           & 77.5           & 73.9           & 77.3           & 80.6           & 71.9           \\
		Co-Mixup~\citep{kim2021co}                & 73.8           & 82.6           & 86.8           & 76.9           & 68.1           & 78.9           & 82.5           & 74.2           \\
		SaliencyMix~\citep{uddin2020saliencymix}  & 75.8           & 79.7           & 82.2           & 84.4           & \se{75.3}      & 81.2           & 83.8           & 79.5           \\
		StyleMix~\citep{hong2021stylemix}         & 73.0           & 74.6           & 72.4           & 73.4           & 72.9           & 79.5           & 78.2           & 74.6           \\
		StyleCutMix~\citep{hong2021stylemix}      & 74.3           & 83.1           & 86.9           & 78.9           & 73.8           & 80.9           & 83.1           & 76.3           \\ \midrule
		AlignMixup (ours)                         & \se{76.1}      & \se{84.3}      & \se{87.1}      & \tb{85.8}      & 74.7           & \se{82.6}      & \se{86.1}      & \se{80.9}      \\
		AlignMixup/AE (ours)                      & \tb{77.0}      & \tb{85.8}      & \tb{87.9}      & \se{83.7}      & \tb{76.2}      & \tb{84.8}      & \tb{87.2}      & \tb{82.3}      \\ \midrule
		Gain                                      & \gp{\tb{+2.1}} & \gp{\tb{+2.7}} & \gp{\tb{+1.0}} & \gp{\tb{+1.4}} & \gp{\tb{+0.9}} & \gp{\tb{+3.6}} & \gp{\tb{+3.4}} & \gp{\tb{+2.8}} \\ \midrule
		\toprule
		\Th{Noise}                                & \mc{4}{\Th{Uniform}}                                              & \mc{4}{\Th{Gaussian}}                                             \\ \midrule
		Baseline                                  & 58.3           & 75.3           & 75.0           & 69.0           & 60.8           & 64.3           & 62.9           & 63.9           \\
		Input~\citep{zhang2018mixup}              & 50.0           & 67.9           & 71.8           & 71.7           & 60.2           & 65.0           & 63.1           & 64.1           \\
		Cutmix~\citep{yun2019cutmix}              & 74.8           & 80.0           & 84.9           & 72.4           & 75.7           & 79.0           & 84.0           & 70.9           \\
		Manifold~\citep{verma2019manifold}        & 69.8           & 75.9           & 83.2           & 71.9           & 70.8           & 78.8           & 81.3           & 71.6           \\
		PuzzleMix~\citep{kim2020puzzle}           & 78.6           & 85.2           & 86.0           & 74.4           & 78.5           & 85.1           & 85.9           & 74.3           \\
		Co-Mixup~\citep{kim2021co}                & 80.4           & 87.6           & 87.4           & 75.2           & 81.6           & 78.6           & 89.5           & 74.2           \\
		SaliencyMix~\citep{uddin2020saliencymix}  & 83.1           & 87.4           & 89.1           & 76.6           & 82.4           & 85.4           & 81.1           & \tb{81.3}      \\
		StyleMix~\citep{hong2021stylemix}         & 75.3           & 71.8           & 77.8           & 65.5           & 78.0           & 75.2           & 84.3           & 71.0           \\
		StyleCutMix~\citep{hong2021stylemix}      & 84.5           & 83.2           & 88.6           & \se{78.3}      & 84.8           & 81.9           & 83.3           & 73.9           \\
		\midrule
		AlignMixup (ours)                         & \se{86.9}      & \se{89.1}      & \se{93.6}      & 77.7           & \tb{86.7}      & \tb{87.9}      & \se{91.8}      & \se{77.4}      \\
		AlignMixup/AE (ours)                      & \tb{88.0}      & \tb{90.6}      & \tb{94.0}      & \tb{80.8}      & \se{86.0}      & \se{87.2}      & \tb{91.9}      & 75.6            \\ \midrule
		Gain                                      & \gp{\tb{+3.5}} & \gp{\tb{+3.0}} & \gp{\tb{+4.9}} & \gp{\tb{+2.5}} & \gp{\tb{+1.9}} & \gp{\tb{+2.8}} & \gp{\tb{+2.4}} & \gn{\tb{-3.9}} \\
		\bottomrule
\end{tabular}
\caption{\emph{Out-of-distribution detection} on different datasets (top) and under different noise (bottom) using PreActResnet18. Det Acc (detection accuracy), AuROC, AuPR (ID) and AuPR (OOD): higher is better. Blue: second best. Gain: increase in performance. TI: TinyImagenet.}
\label{tab:ood_extra}
\end{table}

\paragraph{ImageNet classification}

Following the training protocol of~\citep{kim2021co}, \autoref{tab:imgnet} reports classification performance when training for 100 epochs on ImageNet. Using the top-1 error (\%) reported for competitors by~\citep{kim2021co}, AlignMixup outperforms all methods, including Co-Mixup~\citep{kim2021co}. Importantly, while the overall improvement by SOTA methods over Baseline is around $1.64\%$, AlignMixup improves SOTA by another $0.4\%$.

\paragraph{Experiments using transformers}
We apply mixup to LeViT-128S~\cite{Graham_2021_ICCV} on ImageNet for 100 epochs. For AlignMixup, we align the feature tensors in the last layer of the convolution stem. The top-1 accuracy is: baseline {67.4\%}, input mixup {68.3\%}, manifold mixup {67.8\%}, CutMix {68.7\%}, AlignMixup {69.9\%}. Thus, we outperform input mixup and CutMix by \tb{1.6\%} and \tb{1.2\%} respectively, which in turn outperform the baseline by \tb{0.9\%} and \tb{1.3\%} respectively. This means that the improvement brought by mixing is roughly doubled.

\paragraph{Out-of-distribution detection}

We compare AlignMixup with SOTA methods, training R-18 on CIFAR-100 as discussed in \autoref{sec:exp_class}. At inference, ID examples are test images from CIFAR-100, while OOD examples are test images from LSUN~\citep{yu2015lsun} and Tiny-ImageNet, resizing OOD examples to $32 \times 32$ to match the resolution of ID images~\citep{yun2019cutmix}. We also use test images from CIFAR-100 with Uniform and Gaussian noise as OOD samples. Uniform is drawn from $\cU(0,1)$ and Gaussian from $\cN(\mu,\sigma)$ with $\mu = \sigma = 0.5$. All SOTA mixup methods are reproduced using the same experimental settings. Following~\citep{hendrycks2016baseline}, we measure \emph{detection accuracy} (Det Acc) using a threshold of $0.5$, \emph{area under ROC curve} (AuROC) and \emph{area under precision-recall curve} (AuPR).

As shown in \autoref{tab:ood_extra}, AlignMixup outperforms SOTA methods under all metrics by a large margin, indicating that it is better in reducing over-confident predictions.

\paragraph{Calibration}

We compare AlignMixup with SOTA methods , training R-18 on CIFAR-100 as discussed in \autoref{sec:exp_class}. All SOTA mixup methods are reproduced using the same experimental settings. We compare qualitatively by plotting accuracy \vs confidence. As shown in \autoref{fig:ece}, while Baseline is clearly overconfident and Input and Manifold mixup are clearly under-confident, AlignMixup results in the best calibration among all competitors. We also compare quantitatively, measuring the \emph{expected calibration error} (ECE)~\citep{guo2017calibration} and \emph{overconfidence error} (OE)~\citep{thulasidasan2019mixup}. As shown in \autoref{tab:supp-calibration}, AlignMixup outperforms SOTA methods by achieveing lower ECE and OE, indicating that it is better calibrated.

\begin{table}
\centering
\footnotesize
\setlength{\tabcolsep}{1.4pt}
\begin{tabular}{lcc} \toprule
	\Th{Metric}                              & \Th{ECE}        & \Th{OE}          \\ \midrule
	Baseline                                 & 10.25           & 1.11             \\
	Input~\citep{zhang2018mixup}             & 18.50           & 1.42             \\
	CutMix~\citep{yun2019cutmix}             & 7.60            & 1.05             \\
	Manifold~\citep{verma2019manifold}       & 18.41           & 0.79             \\
	PuzzleMix~\citep{kim2020puzzle}          & 8.22            & 0.61             \\
	Co-Mixup~\citep{kim2021co}               & 5.83            & 0.55             \\
	SaliencyMix~\citep{uddin2020saliencymix} & 5.89            & 0.59             \\
	StyleMix~\citep{hong2021stylemix}        & 11.43           & 1.31             \\
	StyleCutMix~\citep{hong2021stylemix}     & 9.30            & 0.87             \\ \midrule
	AlignMixup (ours)                        & \se{5.78}       & \tb{0.41}        \\
	AlignMixup/AE (ours)                     & \tb{5.06}       & \se{0.48}        \\ \midrule
	Gain                                     & \gp{\tb{+0.77}} & \gp{\tb{+0.14}}  \\ \bottomrule
\end{tabular}
\caption{\emph{Calibration} using PreActResnet18 on CIFAR-100. \\ECE: expected calibration error; OE: overconfidence error. Lower is better. Blue: second best. Gain: reduction of error.}
\label{tab:supp-calibration}
\end{table}

\paragraph{Qualitative results of WSOL}

\begin{figure}[t]
\centering
\small
\setlength{\tabcolsep}{1.4pt}
\renewcommand{\arraystretch}{.6}
\newcommand{\hei}{26pt}
\newcommand{\wsol}[4]{%
	\rotatebox[origin=c]{90}{\parbox[c]{\hei}{\centering \footnotesize #4}} &
	\fig{wsol/#1/img1} &
	\fig{wsol/#1/cam_img1} &
	\fig{wsol/#1/img2} &
	\fig{wsol/#1/cam_img2} \\
	& {\scriptsize IoU = #2} &
	& {\scriptsize IoU = #3} & \\[3pt]
}
\begin{tabular}{
	>{\centering}m{16pt}
	>{\centering}m{52pt}
	>{\centering}m{52pt}
	>{\centering}m{52pt}
	>{\centering\arraybackslash}m{52pt}
}
	\wsol{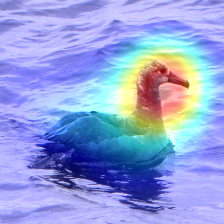}{0.27}{0.41}{Input mixup~\cite{zhang2018mixup}}
	\wsol{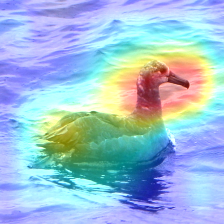}{0.59}{0.52}{CutMix~\cite{yun2019cutmix}}
	\wsol{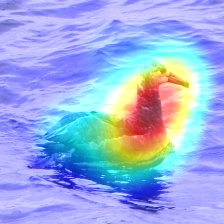}{0.76}{0.63}{AlignMixup (Ours)}
\end{tabular}
\caption{\emph{Localization examples} using ResNet-50 on CUB200-2011. Red boxes: predicted; green: ground truth. }
\label{fig:wsol}
\end{figure}

Qualitative localization results shown in~\autoref{fig:wsol} indicate that
AlignMixup encodes semantically discriminative representations, resulting in better localization performance.

\paragraph{Object detection}

Following the settings of CutMix~\cite{yun2019cutmix}, we use Resnet-50 pretrained on ImageNet using AlignMixup as the backbone of SSD~\cite{liu2016ssd} and Faster R-CNN~\cite{ren2015faster} detectors and fine-tune it on Pascal VOC07~\cite{everingham2010pascal} and MS-COCO~\cite{lin2014microsoft} respectively. AlignMixup outperforms CutMix mAP by \tb{0.8\%} (77.6 $\rightarrow$ 78.4) on Pascal VOC07 and \tb{0.7\%} (35.16 $\rightarrow$ 35.84) on MS-COCO.

\section{Additional ablations}
\label{sec:supp-add-ablation}

\begin{table}[h]
\centering
\footnotesize
\setlength{\tabcolsep}{2pt}
\begin{tabular}{lcccccccc} \toprule
	\Th{Iterations ($i$)}    & 0       & 10      & 20      & 50     & 100     & 200    & 500     & 1000  \\ \midrule
	AlignMixup               & 80.98   & 80.96   & 81.31   & 81.42  & 81.71   & 81.50  & 81.34   & 81.28 \\ \bottomrule
\end{tabular}
\caption{\emph{Ablation} of the number of iterations in Sinkhorn-Knopp algorithm using R-18 on CIFAR-100. Top-1 classification accuracy(\%): higher is better. }
\label{tab:iterations}
\end{table}

\paragraph{Iterations in Sinkhorn-Knopp}

The default number of iterations for the Sinkhorn-Knopp algorithm in solving~\eq{opt} is $i=100$. Here, we investigate more choices, as shown in~\autoref{tab:iterations}. The case of $i=0$ is similar to cross-attention. In this case, we only normalize either the rows or columns in~\eq{poly} once, such that $P \vone = \vone/r$ (when $\vA$ aligned to $\vA'$) or $P\tran \vone = \vone/r$ (when $\vA'$ aligned to $\vA$). We observe that while AlignMixup outperforms the best baseline--StyleCutMix (80.66)--in all cases, it performs best for $i = 100$ iterations.


\end{document}